\documentclass{article}
\usepackage{iclr2027_conference,times}
\iclrfinalcopy 
\usepackage{amsmath,amssymb,amsthm,amsfonts,mathtools,bm}
\usepackage{booktabs,array,multirow,graphicx,microtype,xcolor}
\usepackage{algorithm,algpseudocode}
\usepackage{enumitem}
\usepackage{hyperref}
\hypersetup{hidelinks,pdfauthor={},pdftitle={Certification Frontiers for Gaussian LoRA}}
\newtheorem{theorem}{Theorem}
\newtheorem{proposition}[theorem]{Proposition}

\newtheorem{corollary}[theorem]{Corollary}
\theoremstyle{definition}
\newcommand{\R}{\mathbb R}
\newcommand{\E}{\mathbb E}
\newcommand{\KL}{\operatorname{KL}}
\newcommand{\klbin}{\operatorname{kl}}
\newcommand{\klinverse}{\operatorname{kl}^{-1}_{+}}
\newcommand{\norm}[1]{\left\lVert#1\right\rVert}
\title{Certification Frontiers for Gaussian LoRA:\\ Independent Priors, Posterior Risk,\\ and Prediction-Preserving Balancing}

\author{\textbf{Joyanta Jyoti Mondal}\textsuperscript{1,*},
\textbf{Ibne Farabi Shihab}\textsuperscript{2,*}
\\
\textsuperscript{1}Department of Computer and Information Sciences, University of Delaware, USA\\
\textsuperscript{2}Department of Computer Science, Iowa State University, USA\\
\small{
\textsuperscript{\textbf{*}}Equal Contribution.
\textbf{Correspondence:}
\href{mailto:ishihab@iastate.edu}{ishihab@iastate.edu}
}
}

\begin{document}
\maketitle
\begin{abstract}
Post-hoc Bayesian fine-tuning places Gaussians around trained low-rank adapters, yet a calibrated posterior does not by itself yield a useful generalization certificate. Such a posterior admits an informative PAC-Bayes certificate only when both the loss of its sampled predictors and its KL divergence from an admissible prior are small. In this research, we characterize this certification frontier for Gaussian LoRA posteriors and separate three interventions: changing the prior, changing the stochastic predictor, and changing only how its complexity is counted. First, an exact isotropic KL envelope eliminates the prior scale and yields a width threshold that excludes posterior widths before sampling, while a zero-KL floor identifies targets that no complexity reduction can reach at a measured risk bound. Second, we minimize KL in closed form over the full $GL(r)$ symmetry of the low-rank factors, leaving every sampled adapter product unchanged, and derive the noncentral objective required when the prior center is trained on an independent split. On a small-pool RoBERTa audit of 567 configurations, the recorded 64-draw summaries imply a certificate floor of $0.7298$ even with zero KL and Chernoff accounting, so reducing complexity alone cannot certify these posteriors at the recorded budgets. For a stable posterior in a controlled Gaussian-factor task, Chernoff accounting certifies risk below $0.1$ on 20 of 20 datasets with 1024 draws, whereas Hoeffding certifies none. On deliberately deformed synthetic rank-four factors, matrix balancing reduces KL by $29.3\%$ on average beyond scalar balancing without changing any prediction. Numerical split-prior scenarios make the remaining risk and complexity budgets explicit.
\end{abstract}

\section{Introduction}
\label{sec:introduction}
A Gaussian posterior over a trained low-rank adapter yields a useful PAC-Bayes certificate only when two quantities are small at once: the loss of the sampled predictors and the posterior's KL divergence from an admissible prior. Neither follows from LoRA's small factor count \citep{hu2021lora} or from the accuracy of the trained checkpoint, because the certificate concerns the loss averaged over sampled weights. We refer to the set of attainable pairs of risk and complexity as the \emph{certification frontier} and study three questions: which posterior widths preserve prediction, which priors make those widths affordable, and which costs can be removed without altering any prediction.

These questions arise directly in post-hoc Bayesian fine-tuning. Laplace-LoRA fits a Gaussian around a trained adapter \citep{yang2024laplacelora}, and Training-Free Bayesianization searches for the largest admissible noise scale around one \citep{shi2025trainingfree}; BLoB instead learns the posterior during fine-tuning \citep{wang2024blob}. All three target calibration, which does not by itself yield a PAC-Bayes certificate. Existing nonvacuous certificates rely on suitable priors, stochastic training, or compression \citep{dziugaite2017computing,dziugaite2021role,dong2025localized,lotfi2024nonvacuous}.

\begin{figure}[t]
\centering
\includegraphics[width=\linewidth]{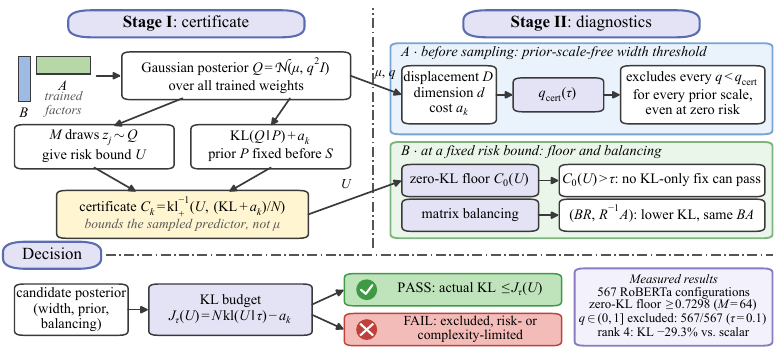}
\caption{Overview. \textbf{Stage I} combines a Monte Carlo risk bound $U$ for the Gaussian posterior $Q$ with $\KL(Q\|P)+a_k$ into the certificate $C_k$. \textbf{Stage II} locates $Q$ on the risk--complexity frontier: (A) a width threshold applied before sampling; (B) at fixed $U$, the zero-KL floor and matrix balancing, which leaves every sampled $BA$ unchanged. \textbf{Decision}: pass when the actual KL fits $J_\tau(U)$.}
\label{fig:overview}
\end{figure}

Our analysis separates what a posterior predicts from how its complexity is counted (Figure~\ref{fig:overview}). We make three contributions:
\begin{itemize}[leftmargin=*,itemsep=1pt,topsep=2pt,parsep=0pt]
\item \textbf{Feasibility tests that need no prior scale.} An exact isotropic KL envelope yields a width threshold that excludes posterior widths before any sampling, and a complexity-free floor identifies targets that no KL correction can reach at the measured risk bound (Section~\ref{sec:frontiers}).
\item \textbf{Prediction-preserving matrix balancing.} We minimize the KL in closed form over the full $GL(r)$ factor symmetry while every sampled product $BA$ stays unchanged, and we derive the noncentral objective required when the prior center is trained on an independent split (Sections~\ref{sec:gauge} and~\ref{sec:split-workflow}).
\item \textbf{Evidence on both sides of the frontier.} On 567 RoBERTa configurations, the recorded 64-draw summaries imply a certificate floor of $0.7298$ even with zero KL and Chernoff accounting. Controlled tests show where tighter sampling helps, and matrix balancing gives a $29.3\%$ mean KL reduction beyond scalar balancing on deformed rank-four factors (Section~\ref{sec:experiments}).
\end{itemize}

%

\section{The stochastic predictor and its certificate}
\label{sec:setup}\label{sec:certificate}
We certify a stochastic predictor that draws its trained weights $z$ from a posterior $Q$. Let $S=((x_{i},y_{i}))_{i=1}^{N}\sim\mathcal {D}^{N}$, with $N\ge2$ and a loss $\ell\in[0,1]$. A configuration $k$ fixes adapter placement, ranks, scaling, prior, and all other trained blocks. LoRA replaces a frozen matrix $W_{0}^{(\ell)}$ by $W_{0}^{(\ell)}+s_{\ell} B^{(\ell)}A^{(\ell)}$ with $\sum_{\ell} r_{\ell}(m_{\ell}+n_{\ell})$ factor entries, but the certified vector $z\in\R^{d_{k}}$ must also include the classification head and every other parameter trained on $S$. Table~\ref{tab:notation} in Appendix~\ref{app:notation} summarizes the notation.

For a posterior $Q$, define
\begin{equation}
R_{\mathcal {D}}(Q)=\E_{z\sim Q}\E_{(x,y)\sim\mathcal {D}}\ell(f_{k,z}(x),y),\qquad
\widehat {R}_{S}(Q)=\E_{z\sim Q}\frac{1}{N}\sum_{i=1}^{N}\ell(f_{k,z}(x_{i}),y_{i}).
\label{eq:risk}
\end{equation}
We refer to the posterior mean $\mu$, which is the trained checkpoint, as the \emph{center}. Its risk $\widehat {R}_{S}(\delta_{\mu})$ can differ substantially from $\widehat {R}_{S}(Q)$; hence adding a Gaussian KL to a center score does not yield a certificate.

\begin{theorem}[PAC-Bayes-kl; \citealp{maurer2004note,alquier2024user}]
\label{thm:pac-bayes-kl}\label{thm:pac-bayes-lora}
For a prior $P$ fixed independently of $S$, with probability at least $1-\delta$ over $S$, simultaneously for all $Q\ll P$,
\begin{equation}
\klbin\!\left(\widehat {R}_{S}(Q)\middle\|R_{\mathcal {D}}(Q)\right)
\le \frac{\KL(Q\|P)+\log(2\sqrt {N}/\delta)}{N}.
\label{eq:pac-bayes-kl}
\end{equation}
\end{theorem}
Here $\klbin$ is binary relative entropy and $\klinverse(u,c)$ is the largest $v\in[u,1]$ such that $\klbin(u\|v)\le c$. A prior may be trained on an independent sample $S_{0}$; conditioning on $S_{0}$ leaves the theorem applicable with denominator $|S|$, not $|S_{0}|+|S|$.

\begin{proposition}[Configuration accounting]
\label{thm:hierarchical}
Let $\mathcal {K}$ be countable and fixed independently of $S$, and let $P_{k}$ and $w_{k}>0$, $\sum_{k} w_{k}=1$, be data-independent priors and weights. On the disjoint union, set $P=\sum_{k} w_{k}\delta_{k}\otimes P_{k}$. Then
\begin{equation}
\KL(\delta_{k}\otimes Q_{k}\|P)=\KL(Q_{k}\|P_{k})+\log(1/w_{k}).
\label{eq:mixture-kl}
\end{equation}
Consequently, the PAC-Bayes event permits selection of $k$ and $Q_{k}$ using $S$.
\end{proposition}
Rank and prior choices are charged through $w_{k}$. A search over posterior widths incurs no additional penalty when $P$ and the hypothesis space are fixed; the reuse of certification draws during that search is addressed by the Monte Carlo bound below.

The empirical risk $\widehat {R}_{S}(Q)$ is estimated by sampling. For each final frozen posterior, draw $z_{1},\ldots,z_{M}\sim Q$ independently and let $Y_{j}=N^{-1}\sum_{i}\ell(f_{z_{j}}(x_{i}),y_{i})$; the independent units are the $M$ draws, not the $MN$ predictions. For $K_{\rm MC}$ posteriors certified together, Hoeffding's inequality gives the simultaneous upper bound
\begin{equation}
U_{H}=\min\left\{1,\widetilde {r}+\sqrt{\frac{\log(K_{\rm MC}/\eta)}{2M}}\right\},\qquad
\widetilde {r}=\frac{1}{M}\sum_{j}Y_{j}.
\label{eq:mc-ucb}
\end{equation}
Because the $Y_{j}$ are bounded (though not Bernoulli), the tighter Chernoff inversion is also valid (Appendix~\ref{app:certificate-proofs}):
\begin{equation}
U_{\rm kl}=\klinverse\!\left(\widetilde {r},\frac{\log(K_{\rm MC}/\eta)}{M}\right).
\label{eq:mc-kl}
\end{equation}
With either upper bound $U$, the certificate is
\begin{equation}
C_{k}=\klinverse\!\left(U_{k},
\frac{\KL(Q_{k}\|P_{k})+\log(1/w_{k})+\log(2\sqrt {N}/\delta_{\rm PB})}{N}\right).
\label{eq:audited-selector}
\end{equation}
It holds with probability at least $1-\delta_{\rm PB}-\eta$ and bounds the risk of the stochastic predictor, not of its center. Minimizing $C_{k}$ over $k$ gives a selector; the accuracy of the center it selects is a separate empirical quantity.

\section{Which isotropic certificates are feasible?}
\label{sec:frontiers}
A useful posterior must preserve prediction while staying close to its prior. For isotropic Gaussians, the complexity side of this tradeoff can be optimized exactly over the prior scale, which yields tests that require no risk estimate.

\begin{theorem}[Oracle isotropic envelope]
\label{thm:isotropic-frontier}
Fix centers $m,\mu\in\R^{d}$, write $D=\norm{\mu-m}_{2}^{2}$, and let $Q_{q}=\mathcal {N}(\mu,q^{2}I)$, $P_{p}=\mathcal {N}(m,p^{2}I)$. Then
\begin{equation}
\inf_{p>0}\KL(Q_{q}\|P_{p})=\frac {d}{2}\log\left(1+\frac{D}{dq^{2}}\right),
\qquad p_{*}^{2}=q^{2}+D/d.
\label{eq:isotropic-envelope}
\end{equation}
For $a_{k}=\log(1/w_{k})+\log(2\sqrt {N}/\delta_{\rm PB})$ and $r(q)=\widehat {R}_{S}(Q_{q})$, the ideal certificate using $r(q)$, and every certificate using an upper bound $U\ge r(q)$, can satisfy $C\le\tau<1$ only if $r(q)\le\tau$ and
\begin{equation}
\frac {d}{2}\log\left(1+\frac{D}{dq^{2}}\right)+a_{k}
\le N\klbin(r(q)\|\tau).
\label{eq:target-feasibility}
\end{equation}
\end{theorem}
Appendix~\ref{app:proof-isotropic} gives the proof.
The optimal scale $p_{*}$ depends on $S$, so it is not an admissible prior. Minimizing over it can only strengthen an exclusion and cannot establish a certificate.

\begin{corollary}[Prior-scale-free width threshold]
\label{cor:isotropic-width}
Suppose $D>0$ and $\tau\in(0,1)$. Define
\begin{equation}
 b_{\tau}=N\log\frac{1}{1-\tau}-a_{k},\qquad
 q_{\rm cert}(\tau)=
 \begin{cases}
 \infty,&b_{\tau}\le0,\\[2pt]
 \displaystyle\sqrt{\frac{D}{d\,[\exp(2b_{\tau}/d)-1]}},&b_{\tau}>0.
 \end{cases}
\label{eq:q-cert}
\end{equation}
Any passing isotropic posterior has $q\ge q_{\rm cert}(\tau)$. Thus $q_{\rm cert}>q_{\max}$ excludes every $q\in(0,q_{\max}]$, for every isotropic prior scale, even if its empirical stochastic risk were zero.
\end{corollary}
The threshold grants zero empirical risk, since $\klbin(r\|\tau)\le\klbin(0\|\tau)$ for $0\le r\le\tau$, so no number of posterior draws can reverse an exclusion. A width above the threshold is only unexcluded and may still destroy prediction. Block-anisotropic posteriors fall outside this result and need their own KL.

\subsection{The best possible accounting-only correction}
\label{sec:accounting-floor}
We next ask whether any reduction in KL could change the decision, holding fixed the predictor, its draw-level losses, the sample size, and the confidence allocation.

\begin{proposition}[Complexity-free certificate floor]
\label{prop:zero-kl-floor}
Let $U\in[0,1]$ be the fixed empirical-risk upper bound and
$a_{k}=\log(1/w_{k})+\log(2\sqrt {N}/\delta_{\rm PB})$. Every value of~\eqref{eq:audited-selector} with a nonnegative KL satisfies
\begin{equation}
 C_{k}\ge C_{0,k}(U):=\klinverse\!\left(U,\frac{a_{k}}{N}\right)\ge U.
 \label{eq:zero-kl-floor}
\end{equation}
If $C_{0,k}(U)>\tau$, no correction that changes only the KL can make this certificate pass $\tau$.
\end{proposition}
The bound holds because $\klinverse(U,c)$ is nondecreasing in $c$. The floor is an arithmetic relaxation, not a certificate: setting the KL to zero does not make $Q=P$. Because the balancing of Section~\ref{sec:gauge} preserves every draw-level loss, the same $U$, and hence the same floor, applies before and after balancing.
The floor therefore separates a removable complexity cost from a binding risk or confidence cost.

\section{Correcting KL without changing a stochastic predictor}
\label{sec:gauge}
Predictions depend only on the product $BA$, but the KL of a factor posterior also depends on how that product is split between the factors. Rescaling only the posterior mean would change the stochastic model, so we map every draw through $(B,A)\mapsto(BR,R^{-1}A)$ and optimize over the full group $GL(r)$ rather than one scalar per adapter.

\begin{theorem}[Exact matrix-gauge envelope]
\label{thm:matrix-gauge}
Let $B\in\R^{m\times r}$ and $A\in\R^{r\times n}$ have a nonsingular joint Gaussian posterior $Q$, and let $P=\mathcal {N}(0,p^{2}I_{(m+n)r})$. Set
\[
 S_{B}=\E_{Q}[B^{\top} B],\qquad S_{A}=\E_{Q}[AA^{\top}],\qquad \Delta=n-m.
\]
Both moment matrices are positive definite. For $R\in GL(r)$, let $Q_{R}=(T_{R})_{\#}Q$ with $T_{R}(B,A)=(BR,R^{-1}A)$ and write $\Gamma=RR^{\top}$. Then
\begin{equation}
\KL(Q_{R}\|P)=C_{Q}+
\frac{\operatorname{tr}(S_{B}\Gamma)+\operatorname{tr}(S_{A}\Gamma^{-1})}{2p^{2}}
+\frac{\Delta}{2}\log\det \Gamma,
\label{eq:matrix-gauge-kl}
\end{equation}
where $C_{Q}$ is independent of $R$. The unique minimizing $\Gamma\succ0$ is
\begin{align}
C&=S_{B}^{1/2}S_{A}S_{B}^{1/2},\qquad
V_{*} =\frac{(p^{4}\Delta^{2}I+4C)^{1/2}-p^{2}\Delta I}{2},\nonumber\\
\Gamma_{*}&=S_{B}^{-1/2}V_{*}S_{B}^{-1/2}.
\label{eq:matrix-gauge-optimum}
\end{align}
Every $R=\Gamma_{*}^{1/2}O$, with $O$ orthogonal, is optimal and induces exactly the same distribution of $BA$ as $Q$.
\end{theorem}
The change of variables contributes $(n-m)\log|\det R|$ to KL; omitting this term gives the wrong optimizer for rectangular factors. Stationarity in $\Gamma$ yields the matrix equation
\begin{equation}
\Gamma S_{B}\Gamma+p^{2}\Delta \Gamma-S_{A}=0.
\label{eq:matrix-gauge-stationarity}
\end{equation}
After congruence by $S_{B}^{1/2}$, it becomes $V^{2}+p^{2}\Delta V=C$. The positive root in each eigendirection gives~\eqref{eq:matrix-gauge-optimum}. Coercivity and uniqueness of this positive solution establish global optimality, without assuming Euclidean convexity of the objective. Appendix~\ref{app:gauge-proof} gives the full proof and a stable implementation.

Balancing leaves posterior risks, center predictions, and every sampled product unchanged; only the KL decreases. Because Theorem~\ref{thm:pac-bayes-kl} holds simultaneously over posteriors, optimizing $R$ costs no extra confidence. It does not, however, repair an unstable posterior. It contains scalar balancing as $\Gamma=c^{2}I$ and applies layer by layer under a factorized prior.


\subsection{Independent nonzero prior centers}
\label{sec:noncentral}
A prior centered at an independent initialization, or at factors learned only on an independent sample $S_{0}$, breaks the rotational symmetry used in Theorem~\ref{thm:matrix-gauge}. Conditioning on all prior-side data and randomness, write $M_{B}=\E_{Q}B$, $M_{A}=\E_{Q}A$, and let
$P_{0}=\mathcal {N}((B_{0},A_{0}),p^{2}I)$ be the fixed prior. A direct change of variables gives
\begin{align}
 \KL(Q_{R}\|P_{0})={}&C_{Q}+(n-m)\log|\det R|\nonumber\\
 &+\frac{\E_{Q}\norm{BR-B_{0}}_{F}^{2}+\E_{Q}\norm{R^{-1}A-A_{0}}_{F}^{2}}{2p^{2}},
 \label{eq:noncentral-gauge}
\end{align}
where $C_{Q}=d\log(2\pi p^{2})/2-h(Q)$ and $d=(m+n)r$. The cross terms depend on $R$ itself, not only on $RR^{\top}$, so the closed form of Theorem~\ref{thm:matrix-gauge} is in general not the noncentral optimizer. In a split-trained prior both $B_{0}$ and $A_{0}$ may be nonzero, and resetting $B_{0}$ to zero after prior training would change the prior.

\begin{proposition}[Exact elimination of the orthogonal factor]
\label{prop:noncentral-polar}
Set $R=\Gamma^{1/2}O$, $\Gamma\succ0$, $O^{\top} O=I$, and
\begin{equation}
 F(\Gamma)=\Gamma^{1/2}M_{B}^{\top} B_{0}+\Gamma^{-1/2}M_{A}A_{0}^{\top}.
 \label{eq:noncentral-polar-matrix}
\end{equation}
With $C_{Q,0}=C_{Q}+(\norm{B_{0}}_{F}^{2}+\norm{A_{0}}_{F}^{2})/(2p^{2})$,
\begin{align}
 \min_{O^{\top} O=I}\KL(Q_{\Gamma^{1/2}O}\|P_{0})
 ={}&C_{Q,0}+\frac{\operatorname{tr}(S_{B}\Gamma)+\operatorname{tr}(S_{A}\Gamma^{-1})}{2p^{2}}\nonumber\\
 &-\frac{\norm{F(\Gamma)}_{*}}{p^{2}}+\frac{n-m}{2}\log\det \Gamma.
 \label{eq:noncentral-reduction}
\end{align}
Here $\norm{\cdot}_{*}$ is the nuclear norm. If $F(\Gamma)=L\Lambda V^{\top}$ is a full singular value decomposition, $O=LV^{\top}$ attains the minimum. A global minimum over $\Gamma\succ0$ exists, but neither uniqueness nor a closed form is asserted.
\end{proposition}
The proof expands~\eqref{eq:noncentral-gauge} and maximizes $\operatorname{tr}(O^{\top} F)$ over orthogonal $O$ (Appendix~\ref{app:noncentral-proof}).
In practice we compare numerical candidates with $R=I$ through the exact KL~\eqref{eq:noncentral-gauge}; any attained reduction is valid without a global optimum.
A better prior center reduces the displacement, but it does not by itself make prediction-preserving widths affordable.

\section{Using the frontier with an independent prior}
\label{sec:split-workflow}
An independent prior split changes which displacement is charged. Let $S_{0}$ be independent of $S$, fit a prior center $\mu_{0}$ using only $S_{0}$, and fix the base model, tokenizer, head, adapter layout, and prior covariance before accessing $S$. Continue learning only the adapter factors on $S$. A head fitted on $S_{0}$ and then frozen is part of the hypothesis; a head fitted on $S$ must be charged even if it is frozen afterwards.

For independent factor blocks with posterior widths $q_{b}$ and fixed prior widths $p_{b}$, evaluate the full Gaussian KL in Equation~\eqref{eq:gaussian-kl}. The displacement is $D=\|\mu-\mu_{0}\|^{2}$, measured in the declared factor coordinates. The identity $\KL=D/(2p^{2})$ applies only when every posterior variance matches the common prior variance. Narrower widths add covariance-mismatch cost.

\subsection{Screening before sampling, deciding after it}
The workflow has three steps. First, once the displacement is known, compute $q_{\rm cert}(\tau)$ and drop the excluded widths; this needs no posterior draws and applies to the raw isotropic family. Second, select candidate widths on pilot draws and freeze them. Third, compute $U$ on an independent final stream and compare the actual KL, after any balancing, with the budget
\begin{equation}
 J_{\tau}(U)=N\klbin(U\|\tau)-a_{k},\qquad U\le\tau.
 \label{eq:measured-kl-budget}
\end{equation}
A candidate passes exactly when its actual KL is at most $J_{\tau}(U)$. If $U>\tau$ or $J_{\tau}(U)<0$, risk or confidence already blocks the certificate. If only the balanced KL fits within the budget, balancing changes the decision without changing any prediction. Appendix~\ref{app:budget-proofs} inverts the budget into sample-size and draw requirements.

These outcomes are decisions rather than learning-curve forecasts. A prospective study should therefore fix its rules before inspecting the final draws and report excluded, unexcluded-but-failing, and passing cases separately.


\subsection{A matched description-length reference}
A compression comparison concerns the decoded predictor, not the Gaussian stochastic predictor. Encode the balanced factor displacement relative to the same $S_{0}$-trained side information using a complete prefix-free code of length $L$ bits, including quantization scales, tensor layout, and every reconstruction choice. Evaluating its actual bounded empirical error $\widehat {r}_{\rm dec}$ gives the standard description-prior comparison
\begin{equation}
 C_{\rm code}=\klinverse\!\left(\widehat {r}_{\rm dec},
 \frac{L\log2+\log(2\sqrt {N}/\delta_{\rm code})}{N}\right).
 \label{eq:code-comparator}
\end{equation}
It follows by giving the codeword prior mass at least $2^{-L}$ in Theorem~\ref{thm:pac-bayes-kl}. This fixed-code reference follows compression bounds \citep{lotfi2024nonvacuous} but does not reproduce SubLoRA's training or loss, and decoded accuracy should be reported beside the code length. Appendix~\ref{app:large-task-protocol} specifies the code and its confidence allocation.

\section{Experiments}
\label{sec:experiments}\label{sec:results}\label{sec:modelselect}
We report three measured studies: a small-pool RoBERTa audit with a conservative reanalysis of its summaries (Section~\ref{sec:selector}), a controlled test of Monte Carlo slack (Section~\ref{sec:mc-exp}), and a controlled test of matrix balancing (Section~\ref{sec:gauge-exp}). Appendix~\ref{app:large-task-protocol} works through numerical split-prior scenarios.

\subsection{Why the small-pool posterior cannot be repaired by accounting}
\label{sec:selector}\label{sec:paper-cert}\label{sec:recorded-accounting}\label{sec:isotropic-screen}
The RoBERTa-base audit uses seven GLUE tasks, requested pools of $256$, $512$, or $1024$ examples, and three seeds. Each of its 63 problems contains nine rank--decay configurations, giving 567 fitted checkpoints. Query/value adapters and the classification head are all trained on the pool, so all are charged. The prior is origin-centered with $p=1/\sqrt{\lambda}$ for LoRA weight decay $\lambda$. Because the historical seven-width search was not fully recorded, we treat this audit as a descriptive study rather than a reproducible validation. Appendix~\ref{app:extended-experiments} gives full settings and selector results.

With eight posterior draws, the Hoeffding half-width of $0.6065$ saturates every risk upper bound at one, although the complete KL is only $2.34$--$45.91$ nats. A matched rerun with 64 draws on 24 problems (216 configurations) gives bounds of $0.736$--$0.928$, but ranking by them selects centers with accuracy $0.7388$, below the $0.7655$ of training-risk selection. Across all 63 problems, the centers chosen by the center-risk proxy, training risk, and retrained validation reach $0.7053$, $0.7066$, and $0.7088$, while their stochastic predictors reach only $0.511$--$0.524$ (Appendix Figure~\ref{fig:selector-certificate}). Accuracy and certification therefore concern different predictors: the deterministic center and its stochastic posterior.

\begin{table}[t]
\centering\small
\caption{Small-pool audit and accounting floors. The last column is the zero-KL floor with Chernoff accounting at the favorable $N=1024$, computed conservatively from rounded summaries; it is a lower bound, not an issued certificate.}
\label{tab:frontier-audit}
\begin{tabular}{rrrrr}
\toprule
Draws & Configurations & Mean KL & Reported certificate & Accounting floor\\
\midrule
8 & 567 & 12.49 & $1$ & $0.9320$\\
64 & 216 & 12.67 & $0.736$--$0.928$ & $0.7298$\\
\bottomrule
\end{tabular}
\end{table}

\begin{figure}[t]
\centering
\includegraphics[width=\linewidth]{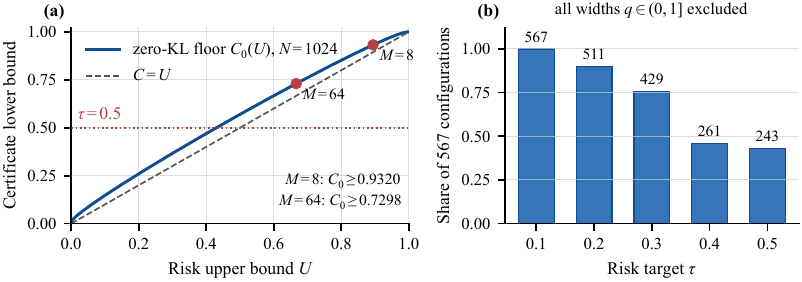}
\caption{Accounting cannot repair the small-pool posteriors. (a) Zero-KL certificate floor $C_{0}(U)$ at the most favorable $N=1024$; the markers are the Chernoff risk floors implied by the recorded 8- and 64-draw summaries, and both floors lie above $\tau=0.5$. (b) Share of the 567 configurations for which the raw isotropic threshold excludes every width $q\in(0,1]$, before any posterior sampling; bar labels are counts.}
\label{fig:floor-screen}
\end{figure}

Even with zero KL and the tighter Chernoff bound, the recorded summaries imply certificates of at least $0.9320$ and $0.7298$ (Table~\ref{tab:frontier-audit}; Figure~\ref{fig:floor-screen}a), so no balancing can rescue these posteriors at the recorded budgets. The floor does not rule out narrower posteriors, other prior centers, or larger samples; Appendix~\ref{app:accounting-floor} gives the arithmetic. Monte Carlo draw budgets lead to the same conclusion. Inverting the Hoeffding budget on the 216 recorded 64-draw configurations, no draw budget suffices for any configuration at $\tau=0.5$; at $\tau=0.7$ none suffices for 28 of them and the median requirement is 343 draws; only at $\tau=0.9$ do 64 draws suffice, for 201 of 216 (Appendix Figure~\ref{fig:draw-budget}).

The width screen reaches a similar conclusion before any sampling. The raw origin-centered isotropic threshold excludes every $q\in(0,1]$ on $243,261,429,511,567$ configurations at targets $0.5,0.4,0.3,0.2,0.1$, respectively (Figure~\ref{fig:floor-screen}b). A nine-width risk sweep cannot sharpen this screen: its simultaneous half-width of $0.913$ makes every risk lower bound zero.

\subsection{Controlled separation of posterior risk and sampling uncertainty}
\label{sec:mc-exp}
To isolate the Monte Carlo mechanism, we evaluate a rank-one Gaussian-factor classifier on a two-dimensional Gaussian classification problem. The center is fitted from $N=2048$ examples, the prior is fixed, and the exact empirical stochastic risk is available analytically. Across 20 independent datasets, a stable posterior with width $q=0.1$ has exact empirical risk $0.0141\pm0.0025$, while $q=4$ gives $0.4756\pm0.0002$ (mean $\pm$ sample standard deviation). All nine widths, four fixed nested draw budgets, and two inequalities are covered by $K_{\rm MC}=72$.

\begin{table}[t]
\centering\small
\caption{Controlled Gaussian-factor classification: mean $\pm$ sample SD of the risk certificate across 20 independent datasets. The same posterior draws are used for both inequalities. The prior, trained center, KL, and confidence allocation are held fixed within each row.}
\label{tab:mc-control}
\begin{tabular}{rrrr}
\toprule
Posterior width $q$ & Draws $M$ & Hoeffding certificate & Chernoff certificate\\
\midrule
0.1 & 8 & $0.777\,\pm\,0.005$ & $0.721\,\pm\,0.008$ \\
0.1 & 64 & $0.326\,\pm\,0.003$ & $0.215\,\pm\,0.007$ \\
0.1 & 256 & $0.189\,\pm\,0.003$ & $0.104\,\pm\,0.006$ \\
0.1 & 1024 & $0.117\,\pm\,0.003$ & $0.064\,\pm\,0.005$ \\
4 & 8 & $0.999\,\pm\,0.005$ & $0.984\,\pm\,0.021$ \\
4 & 64 & $0.804\,\pm\,0.038$ & $0.792\,\pm\,0.033$ \\
4 & 256 & $0.684\,\pm\,0.016$ & $0.683\,\pm\,0.016$ \\
4 & 1024 & $0.627\,\pm\,0.009$ & $0.627\,\pm\,0.009$ \\
\bottomrule
\end{tabular}
\end{table}

\begin{figure}[t]
\centering
\includegraphics[width=\linewidth]{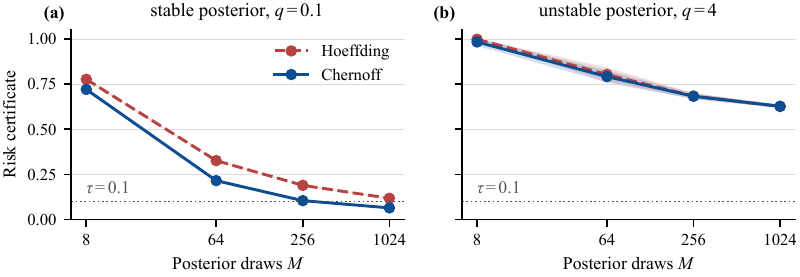}
\caption{Posterior behavior and Monte Carlo slack have different effects. (a) The stable posterior ($q=0.1$) benefits from more draws and from the Chernoff inequality, which crosses $\tau=0.1$ at $M=1024$. (b) The unstable posterior ($q=4$) stays far above the target under both inequalities. Lines are means and bands $\pm1$ sample SD over 20 datasets (Table~\ref{tab:mc-control}); both inequalities use the same posterior draws and simultaneous confidence family.}
\label{fig:mc-controls}
\end{figure}

At $q=0.1$ and $M=1024$, Chernoff certifies risk below $0.1$ on $20/20$ datasets, compared with $0/20$ for Hoeffding. The unstable posterior remains far above that target under either inequality (Table~\ref{tab:mc-control}; Figure~\ref{fig:mc-controls}). Saturation at eight draws therefore reflects insufficient sampling rather than an impossibility for the posterior family. This positive control validates the implementation on a tractable problem; it does not show that more draws alone would certify the language-model posteriors.

\subsection{Controlled prediction-preserving matrix balancing}
\label{sec:gauge-exp}
\begin{figure}[t]
\centering
\includegraphics[width=\linewidth]{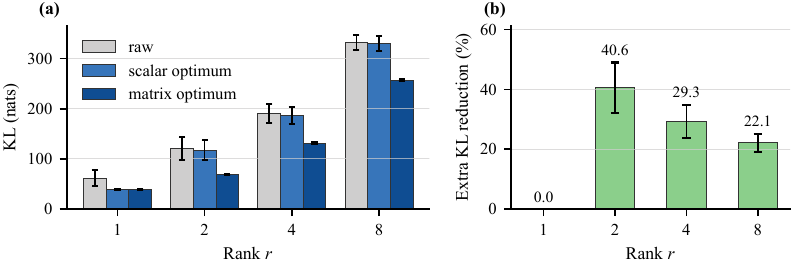}
\caption{Prediction-preserving balancing on deformed Gaussian factors ($m=12$, $n=8$; 20 seeds, error bars $\pm1$ sample SD). (a) Raw, scalar-balanced, and matrix-balanced KL describe the same stochastic product. (b) Mean per-case KL reduction of the matrix optimum beyond scalar balancing; the two optima coincide at rank one.}
\label{fig:gauge}
\end{figure}

This experiment uses synthetic anisotropic factor posteriors. We generate Gaussian factor posteriors for three rectangular or square shapes, four ranks, and 20 seeds, yielding $240$ cases. For each posterior, we compare its raw KL, the exact scalar optimum, and Theorem~\ref{thm:matrix-gauge}; all use the same fixed prior and stochastic product distribution. At $m=12,n=8,r=4$, full balancing reduces KL from the scalar optimum $186.95\pm16.82$ to $131.21\pm1.55$ nats (Table~\ref{tab:gauge-results}; Figure~\ref{fig:gauge}). Sampled predictions are unchanged by construction, so the entire reduction occurs in the complexity term.

\begin{table}[t]
\centering\small\setlength{\tabcolsep}{4pt}
\caption{Gaussian factor experiment, $m=12,n=8$. KL is in nats; entries are mean $\pm$ sample SD over 20 seeds. The last column is the mean per-case percentage reduction relative to scalar balancing, not a ratio of pooled means. All three columns describe the same stochastic product.}
\label{tab:gauge-results}
\begin{tabular}{rrrrr}
\toprule
Rank & Raw KL & Scalar optimum & Matrix optimum & Extra reduction (\%)\\
\midrule
1 & $61.75\,\pm\,16.60$ & $39.04\,\pm\,1.26$ & $39.04\,\pm\,1.26$ & $0.0\,\pm\,0.0$ \\
2 & $120.40\,\pm\,22.29$ & $117.59\,\pm\,19.87$ & $68.35\,\pm\,1.31$ & $40.6\,\pm\,8.4$ \\
4 & $190.76\,\pm\,18.60$ & $186.95\,\pm\,16.82$ & $131.21\,\pm\,1.55$ & $29.3\,\pm\,5.6$ \\
8 & $332.50\,\pm\,15.37$ & $329.86\,\pm\,15.22$ & $256.67\,\pm\,2.33$ & $22.1\,\pm\,3.1$ \\
\bottomrule
\end{tabular}
\end{table}

Across all shapes and ranks, the largest relative stationarity residual is $1.06\times10^{-13}$, the largest relative disagreement with independent numerical minimization is $9.91\times10^{-16}$, and paired sampled products differ by at most $2.71\times10^{-15}$. The scalar and matrix optima coincide at rank one, as required. These measurements validate the closed form.
Because the deformation is deliberate, these percentages show that balancing can reduce KL substantially; they do not estimate the reduction for trained LoRA factors. Its effect on a given posterior depends on whether that posterior's zero-KL floor lies below the target, which is not the case in the small-pool audit (Section~\ref{sec:selector}).

\section{Related work}
\label{sec:related}
Post-hoc uncertainty and risk certification have different objectives. Laplace-LoRA, Training-Free Bayesianization, and BLoB build low-rank Gaussian posteriors for calibration \citep{yang2024laplacelora,shi2025trainingfree,wang2024blob}; we ask whether such a posterior also affords a bounded-loss certificate, without substituting calibration quality for empirical Gibbs risk.

Nonvacuous neural-network bounds optimize posterior risk and complexity together \citep{dziugaite2017computing,pitas2020meanfield}. Prior-data allocation and curvature affect that tradeoff \citep{dziugaite2021role,yang2022capacity}; localized priors, PAC-tuning, and randomized LoRA modify training or the certificate construction \citep{dong2025localized,liu2023pactuning,lei2024rlora}. Our independent split-trained prior is not the same-sample localized objective. Compression-based language-model bounds \citep{lotfi2024nonvacuous} provide a description-length reference but certify a different decoded predictor and loss construction. The matrix calculation specializes the established role of rescaling invariance \citep{rouchouse2025rescaling}; it does not introduce invariance as a new principle. Appendix~\ref{app:extended-related} reviews further work on PAC-Bayes certificates, Bayesian LoRA, low-rank adaptation theory, parameter symmetries, and derandomization.

\section{Limitations}
\label{sec:limitations}
Our language-model evidence is a small-pool audit (RoBERTa-base, seven GLUE tasks, pools of at most 1024 examples, an origin-centered prior) whose historical width search was not fully recorded. Its zero-KL floors use the recorded 8- and 64-draw summaries at the favorable $N=1024$; more draws, narrower posteriors, or other prior centers can move them. The width threshold covers only raw isotropic posteriors, the balancing gains come from deliberately deformed synthetic factors, and the noncentral objective has no global-optimality guarantee. The split-prior study and the description-length comparison are specified with numerical scenarios but not yet measured on large tasks.

\section{Discussion and Conclusion}
\label{sec:conclusion}
We study when a Gaussian distribution around a trained low-rank adapter supports a useful PAC-Bayes certificate, and answer with three tools that locate a posterior on the risk--complexity frontier (Figure~\ref{fig:overview}): a prior-scale-free width threshold that excludes isotropic widths before sampling, a zero-KL floor that decides whether any KL reduction can change the outcome at a measured risk bound, and closed-form matrix balancing over $GL(r)$ that leaves every sampled product unchanged, with a noncentral objective for independently trained priors. On the small-pool audit, the recorded summaries alone show that the certificate is limited by risk and confidence rather than complexity: even at zero KL it stays at or above $0.7298$, so improvement requires a lower risk bound or a larger certification sample rather than further accounting. The controlled studies show that the other levers can be effective when their conditions hold: with 1024 draws, Chernoff accounting certifies a stable posterior below $0.1$ on all 20 datasets, and balancing reduces KL by a further $29.3\%$ on synthetic rank-four factors, a closed-form instance of the rescaling invariance studied by \citet{rouchouse2025rescaling}. In practice, a certification pipeline can compute the threshold and floor before sampling, balance only when the floor lies below the target, and certify the stochastic predictor itself. Whether these tools yield nonvacuous certificates for trained adapters remains to be measured on large tasks with independently trained priors.

\section*{Reproducibility statement}
The appendix contains complete proofs, the measured tables, and the specified inputs of every scenario. The accompanying scenario file records assumed losses and complexities separately from their formula-derived certificates, thresholds, margins, and workload counts.
\section*{AI use statement}
Generative AI assisted with literature discovery, mathematical exposition, code preparation, numerical analysis, and language editing. The authors remain responsible for the claims, experimental interpretation, and final submission.
\section*{Ethics statement}
The study uses established language-understanding benchmarks and synthetic numerical examples. It collects no new human-subject data. A bound for a specified bounded loss is not a general safety guarantee.
\bibliographystyle{iclr2027_conference}
\bibliography{references}
\clearpage
\appendix

\section{Notation}
\label{app:notation}
Table~\ref{tab:notation} collects the symbols used in the main text, grouped by topic. Symbols used only inside a single proof are defined where they appear.

\begin{table}[ht]
\centering\small
\caption{Notation.}
\label{tab:notation}
\begin{tabular}{p{.30\linewidth}p{.64\linewidth}}
\toprule
Symbol & Meaning\\
\midrule
\multicolumn{2}{l}{\textit{Data, predictor, and certificate}}\\
$S$, $N$, $\mathcal{D}$ & certification sample, its size, and the data distribution\\
$\ell\in[0,1]$ & bounded loss\\
$k\in\mathcal{K}$, $w_{k}$ & configuration and its prior weight, $\sum_{k}w_{k}=1$\\
$W_{0}^{(\ell)}$, $B^{(\ell)}$, $A^{(\ell)}$, $s_{\ell}$ & frozen matrix, LoRA factors, and scaling of adapted layer $\ell$\\
$r$, $m$, $n$ & adapter rank and factor dimensions, $B\in\R^{m\times r}$, $A\in\R^{r\times n}$\\
$z\in\R^{d_{k}}$ & all trained coordinates (adapters, head, and other trained blocks)\\
$P$, $Q$ ($P_{k}$, $Q_{k}$) & prior and posterior over $z$ (for configuration $k$)\\
$\mu$ & posterior mean, the trained checkpoint (the \emph{center})\\
$R_{\mathcal{D}}(Q)$, $\widehat{R}_{S}(Q)$ & population and empirical risk of the stochastic predictor\\
$\klbin$, $\klinverse(u,c)$ & binary KL and its upper inversion\\
$M$, $z_{j}$, $Y_{j}$, $\widetilde{r}$ & posterior draws, a draw, its empirical loss, and their mean\\
$K_{\rm MC}$ & number of posteriors covered by one Monte Carlo family\\
$U_{H}$, $U_{\rm kl}$, $U$ & Hoeffding, Chernoff, and generic upper bounds on $\widehat{R}_{S}(Q)$\\
$\delta_{\rm PB}$, $\eta$ & PAC-Bayes and Monte Carlo failure probabilities\\
$C_{k}$ & certificate of configuration $k$\\
\midrule
\multicolumn{2}{l}{\textit{Isotropic feasibility}}\\
$q$, $p$, $p_{*}$ & posterior width, prior width, and KL-optimal prior width\\
$m$, $D$, $d$ & prior center, displacement $D=\norm{\mu-m}_{2}^{2}$, and dimension\\
$a_{k}$ & confidence cost $\log(1/w_{k})+\log(2\sqrt{N}/\delta_{\rm PB})$\\
$\tau$ & target risk\\
$b_{\tau}$, $q_{\rm cert}(\tau)$ & available complexity at zero risk and the width threshold\\
$C_{0,k}(U)$ & zero-KL (complexity-free) certificate floor\\
\midrule
\multicolumn{2}{l}{\textit{Matrix balancing}}\\
$R\in GL(r)$, $T_{R}$, $Q_{R}$ & gauge matrix, map $(B,A)\mapsto(BR,R^{-1}A)$, and pushforward of $Q$\\
$\Gamma=RR^{\top}$, $\Gamma_{*}$ & positive-definite gauge variable and its optimum\\
$S_{B}$, $S_{A}$ & second moments $\E_{Q}[B^{\top}B]$ and $\E_{Q}[AA^{\top}]$\\
$\Delta$ & dimension gap $n-m$\\
$C$, $V_{*}$ & $S_{B}^{1/2}S_{A}S_{B}^{1/2}$ and the root defining $\Gamma_{*}$\\
$C_{Q}$ & $R$-independent constant of the KL, with entropy $h(Q)$\\
$S_{0}$, $\mu_{0}$ & independent prior sample and prior center fitted on it\\
$B_{0}$, $A_{0}$, $P_{0}$ & prior factor centers and the noncentral prior\\
$M_{B}$, $M_{A}$, $F(\Gamma)$, $C_{Q,0}$ & posterior factor means, orientation matrix, and noncentral constant\\
\midrule
\multicolumn{2}{l}{\textit{Workflow and experiments}}\\
$J_{\tau}(U)$ & KL budget $N\klbin(U\|\tau)-a_{k}$\\
$L$, $\widehat{r}_{\rm dec}$, $\delta_{\rm code}$, $C_{\rm code}$ & code length in bits, decoded error, its failure probability, and certificate\\
$\lambda$ & LoRA weight decay; the audit prior has $p=1/\sqrt{\lambda}$\\
\bottomrule
\end{tabular}
\end{table}

\section{Extended related work}
\label{app:extended-related}
\paragraph{PAC-Bayes foundations.} The certificate of Theorem~\ref{thm:pac-bayes-kl} belongs to the PAC-Bayesian line, in which the empirical term averages the loss over the posterior, complexity is measured against a data-independent prior, and the guarantee holds simultaneously over posteriors \citep{mcallester1999pac,langford2002pac,seeger2002pac,maurer2004note,catoni2007pac,alquier2024user}. Data-dependent priors require a separate argument, such as differential privacy or an independent data split \citep{dziugaite2018datadependent,dziugaite2021role}.

\paragraph{Tighter and nonvacuous certificates.} Following the first nonvacuous bounds for stochastic networks \citep{dziugaite2017computing}, later work tightens the risk--complexity pair through training objectives \citep{perez2021tighter}, compression \citep{zhou2018non,lotfi2022tight}, recursive or meta-learned priors \citep{wu2024recursive,leblanc2025metalearned}, sharper kl inversions \citep{garciaperez2025tightness}, tail-robust and complexity-measure bounds \citep{rodriguezgalvez2023more,viallard2024leveraging}, certifiable surrogates \citep{bazinet2026bound}, and task-specific certificates \citep{vanelst2025contrastive}. Bounds for unmodified trained networks \citep{than2025nonvacuous} and an assessment of how well PAC-Bayes explains generalization \citep{germain2025howgood} are closest in spirit to our post-hoc setting.

\paragraph{Certificates for language models and post-training.} Compression bounds for language models trained from scratch \citep{lotfi2024nonvacuous} extend to larger models by treating tokens as data points \citep{lotfi2024unlocking}, and quantized adapters yield certificates for language-model safety monitors \citep{lamb2026tight}. Black-box post-training, model merging, prompt optimization, and reinforcement learning with verifiable rewards also admit nonvacuous bounds \citep{labiad2025tuning,kim2025merging,madras2025prompts,zhu2026rlvr}. These methods train or encode for the bound, whereas we evaluate a Gaussian attached after ordinary fine-tuning.

\paragraph{Bayesian and uncertainty-aware LoRA.} Beyond the methods discussed in Section~\ref{sec:related}, variational, subspace, weight-averaging, ensemble, and orthogonalized constructions target uncertainty and calibration \citep{lin2026bayesianlora,cong2025ivon,marszalek2025minimal,huang2026polar,onal2024gswa,samplawski2025svs,coscia2026balora,balabanov2024ensembles}. Calibration gains do not by themselves imply PAC-Bayes validity, although noise-stable posteriors may improve the empirical-risk side of the frontier.

\paragraph{Theory of low-rank adaptation.} Kernel, neural-tangent, expressivity, convergence, and sample-complexity analyses ask whether low-rank adaptation can represent, reach, or learn a solution \citep{malladi2023kernel,jang2024ntk,zeng2024expressive,sun2026convergent,arunan2026sample}, and randomized asymmetric adapters with one frozen factor admit sharp generalization bounds \citep{kratsios2025sharp}. Studies of learning rates, forgetting, and sharpness-aware training show that rank alone does not determine adaptation behavior \citep{hayou2024lora,biderman2024lora,liu2026bilora}. None of these analyses certifies a declared stochastic posterior.

\paragraph{Symmetry, curvature, and posterior geometry.} Rescaling invariances and continuous symmetries make weight-space KL depend on the parameterization, and recent work moves the analysis to quotient or behavioral spaces \citep{rouchouse2025rescaling,aladrah2026quotient,honavar2026behavioral}. Low-rank Bayesian networks induce singular posteriors on the rank-$r$ product manifold \citep{toure2026singular}, and curvature-, norm-, and sharpness-based analyses relate posterior or loss geometry to generalization \citep{pitas2020meanfield,yang2022capacity,yi2026unified,ahmed2026fisher}. Our matrix balancing resolves the $GL(r)$ symmetry of Gaussian factor posteriors in closed form under a fixed isotropic prior.

\paragraph{From stochastic to deterministic predictors.} Transferring a stochastic bound to a deterministic checkpoint requires margin, smoothness, or stability arguments \citep{bartlett2017spectrally,neyshabur2018pac,biggs2022margins,paquin2026smooth,li2026fragileflow}; for low-rank Bayesian networks, balanced factors resolve the gauge ambiguity that such a transfer faces \citep{toure2026deterministic}. We certify and deploy the stochastic predictor and therefore require no such transfer.

\paragraph{Gibbs posteriors and Langevin dynamics.} An alternative to the Gaussian route certifies the Gibbs posterior targeted by regularized stochastic training. Its KL admits a path-sampling identity \citep{gelman1998simulating,lefebvre2010path} that functional inequalities can bound \citep{brascamp1976extensions,otto2000generalization,bobkov2009weighted,bakry2014analysis,cattiaux2014semi}, and Langevin convergence results relate such posteriors to stochastic training \citep{raginsky2017non,vempala2019rapid,erdogdu2022convergence}. We do not pursue this route, because a numerically useful certificate would require these constants to be controlled along the entire temperature path.

\section{Certificate proofs and boundary cases}
\label{app:certificate-proofs}\label{app:assumption-roles}
\subsection{Gaussian KL and configuration selection}
For independent Gaussian blocks $Q_{b}=\mathcal {N}(\mu_{b},q_{b}^{2}I_{d_{b}})$ and $P_{b}=\mathcal {N}(m_{b},p_{b}^{2}I_{d_{b}})$, direct evaluation of the log-density ratio gives
\begin{equation}
\KL(Q\|P)=\frac{1}{2}\sum_{b}\left[
\frac{\norm{\mu_{b}-m_{b}}_{2}^{2}}{p_{b}^{2}}
+d_{b}\left(\frac{q_{b}^{2}}{p_{b}^{2}}-1-\log\frac{q_{b}^{2}}{p_{b}^{2}}\right)\right].
\label{eq:gaussian-kl}
\end{equation}
Every block contributing to the trained predictor must appear. The theorem does not require its optimizer's regularizer to match the prior, so a head trained without weight decay can still be charged to a fixed Gaussian prior. It does not, however, acquire a Gibbs or maximum-a-posteriori interpretation from that accounting choice.

On component $k$ of the disjoint union, the density ratio is
\[
\frac{d(\delta_{k}\otimes Q_{k})}{dP}(k,z)=\frac{1}{w_{k}}\frac{dQ_{k}}{dP_{k}}(z).
\]
Its posterior expectation gives~\eqref{eq:mixture-kl}. Applying Theorem~\ref{thm:pac-bayes-kl} on the union proves simultaneous validity after posterior and configuration selection. A data-dependent prior requires separate accounting; posterior-side optimization does not turn a data-dependent prior into an independent one.

\subsection{Monte Carlo upper bounds}
Conditional on $S$ and a frozen $Q_{k}$, the variables $Y_{k,j}$ are independent across posterior draws, bounded in $[0,1]$, and have mean $\widehat {R}_{S}(Q_{k})$. Hoeffding's inequality gives
\[
\Pr\{\widehat {R}_{S}(Q_{k})>\widetilde {r}_{k}+e\mid S,Q_{k}\}\le e^{-2Me^{2}}.
\]
Taking $e=\sqrt{\log(K_{\rm MC}/\eta)/(2M)}$ and union-bounding over the fixed family proves~\eqref{eq:mc-ucb}. Dependence between different candidate streams is allowed; independence within a posterior's draw sequence is required.

For the Chernoff form, convexity on $[0,1]$ gives $\E e^{tY}\le1-r+re^{t}$ when $\E Y=r$. For $v<r$, exponential Markov with $t<0$ and minimization over $t$ yield
\[
\Pr\{\widetilde {r}\le v\}\le\exp[-M\klbin(v\|r)].
\]
Inverting at level $\eta/K_{\rm MC}$ proves~\eqref{eq:mc-kl}. This argument does not assume that a draw-level average is Bernoulli. Intersecting the Monte Carlo event with the PAC-Bayes event gives total failure probability at most $\delta_{\rm PB}+\eta$. When several budgets or inequality choices are inspected, their union count must also be included. The controlled experiment pays for both; the original fixed-budget experiment uses its separately stated count.

\subsection{Proof of the isotropic envelope}
\label{app:proof-isotropic}
For $x=p^{2}>0$, the Gaussian expression is
\[
K(x,q)=\frac{1}{2}\left[\frac{D+dq^{2}}{x}-d+d\log\frac{x}{q^{2}}\right].
\]
Its derivative is $\tfrac{1}{2}[-(D+dq^{2})/x^{2}+d/x]$, so the only stationary point is $x_{*}=q^{2}+D/d$. The objective diverges as $x\downarrow0$ or $x\uparrow\infty$, making this the global minimum. Substitution gives~\eqref{eq:isotropic-envelope}.

For fixed $r\le\tau$, binary KL is increasing in its second argument above $r$. Hence $\klinverse(r,c)\le\tau$ is equivalent to $c\le\klbin(r\|\tau)$. The Monte Carlo upper bound is no smaller than $r$ on its validity event, and replacing that upper bound by the true empirical posterior risk only relaxes feasibility. Replacing the KL by its oracle minimum relaxes it further, yielding~\eqref{eq:target-feasibility}. An oracle scale depending on $S$ cannot support a positive claim under the independent-prior theorem.

\subsection{Proof of the width threshold}
For $0<r\le\tau$,
\[
\frac{\partial}{\partial r}\klbin(r\|\tau)=\log\frac{r(1-\tau)}{\tau(1-r)}\le0.
\]
Continuity at $r=0$ yields $\klbin(r\|\tau)\le\log(1/(1-\tau))$. Thus any passing width satisfies $\tfrac {d}{2}\log(1+D/(dq^{2}))\le b_{\tau}$. For $D>0$, its left side is strictly positive at every finite width. If $b_{\tau}\le0$, none is feasible; otherwise exponentiation gives~\eqref{eq:q-cert}. Monotonicity in $q$ makes exclusion over $(0,q_{\max}]$ exact. If $D=0$, the envelope is zero and no positive width threshold follows: the necessary condition is $a_{k}\le N\log(1/(1-\tau))$. The implementation uses \texttt{expm1} for the denominator when its argument is small.

\subsection{Data and posterior-draw budgets}
\label{sec:budget-selection}
Fix $U=u<\tau$ and $\kappa=\KL(Q\|P)+\log(1/w_{k})$. Writing $b=\klbin(u\|\tau)>0$, inversion gives the exact arithmetic condition
\begin{equation}
C\le\tau\quad\Longleftrightarrow\quad
Nb-\tfrac{1}{2}\log N\ge\kappa+\log(2/\delta_{\rm PB}).
\label{eq:sample-budget}
\end{equation}
The upper root yields a sample-size threshold at fixed $(u,\kappa)$. It is not a learning-curve prediction because both quantities may change with more training data. No positive-cost certificate passes when $u\ge\tau$.

At fixed $N$ and $\kappa$, let $u_{\tau}\in[0,\tau)$ solve $\klbin(u_{\tau}\|\tau)=(\kappa+\log(2\sqrt {N}/\delta_{\rm PB}))/N$. A solution exists only if that cost is at most $\klbin(0\|\tau)$. Conditional on an observed mean $\widetilde {r}<u_{\tau}$, the Hoeffding implementation passes exactly when
\begin{equation}
M\ge\frac{\log(K_{\rm MC}/\eta)}{2(u_{\tau}-\widetilde {r})^{2}}.
\label{eq:mc-budget}
\end{equation}
These are two different resource limitations. Insufficient posterior draws enlarge $U$; an excessive KL cannot be repaired by Monte Carlo precision. The budgets must be fixed beforehand or covered by a simultaneous or sequentially valid construction.

\subsection{Budget inversion proofs}
\label{app:budget-proofs}
For $u<\tau$, rearranging the certificate condition gives~\eqref{eq:sample-budget}. The function $f(N)=Nb-\tfrac{1}{2}\log N$ decreases to its unique minimum at $1/(2b)$ and increases thereafter. Its upper crossing gives the eventual passing threshold; checking integers directly handles a crossing outside the admissible range $N\ge2$. This arithmetic calculation conditions on fixed $u$ and $\kappa$ rather than assuming that learning leaves them unchanged.

The map $u\mapsto\klbin(u\|\tau)$ is continuous and decreasing on $[0,\tau]$, from $\klbin(0\|\tau)$ to zero. This gives the stated existence and uniqueness of $u_{\tau}$ for positive complexity cost. Substituting $U_{H}$ and solving $U_{H}\le u_{\tau}$ proves~\eqref{eq:mc-budget}. At the boundary $u_{\tau}=0$, no finite Hoeffding budget passes with a nonnegative observed mean. Repeatedly inspecting the bound and stopping on success is not justified by a fixed-$M$ inequality.

\section{Matrix and scalar gauge proofs}
\label{app:gauge-proof}
\subsection{Density transformation and global optimum}
Write $d=(m+n)r$. Since $T_{R}$ acts independently on each of the $m$ rows of $B$ and $n$ columns of $A$, its absolute Jacobian determinant is $|\det R|^{m-n}$. Differential entropy therefore transforms as
\[
h(Q_{R})=h(Q)+(m-n)\log|\det R|.
\]
The Gaussian prior's negative log density is $\tfrac {d}{2}\log(2\pi p^{2})+(\norm {B}_{F}^{2}+\norm {A}_{F}^{2})/(2p^{2})$. Taking expectation under the pushed-forward posterior, and writing $\Gamma=RR^{\top}$, gives
\[
\E\norm{BR}_{F}^{2}=\operatorname{tr}(S_{B}\Gamma),\qquad
\E\norm{R^{-1}A}_{F}^{2}=\operatorname{tr}(S_{A}\Gamma^{-1}).
\]
Combining these expressions proves~\eqref{eq:matrix-gauge-kl}, with $C_{Q}=\tfrac {d}{2}\log(2\pi p^{2})-h(Q)$. Cross-covariances between $B$ and $A$ do not alter these two second-moment terms or the Jacobian calculation. For the independent isotropic blocks used in the experiment,
\[
C_{Q}=-\frac {d}{2}+\frac {d}{2}\log p^{2}-\frac{1}{2}\left(mr\log s_{B}^{2}+nr\log s_{A}^{2}\right).
\]
Nonsingularity makes $S_{B},S_{A}$ positive definite: for any nonzero $v$, $\E\norm{Bv}^{2}$ and $\E\norm{v^{\top} A}^{2}$ are strictly positive.

Differentiating over symmetric positive-definite $\Gamma$ gives
\[
\nabla_{\Gamma}\KL=\frac{1}{2p^{2}}(S_{B}-\Gamma^{-1}S_{A}\Gamma^{-1})+\frac{\Delta}{2}\Gamma^{-1}.
\]
Multiplying stationarity on both sides by $\Gamma$ gives~\eqref{eq:matrix-gauge-stationarity}. Set $V=S_{B}^{1/2}\Gamma S_{B}^{1/2}$. Then $V^{2}+p^{2}\Delta V=C$. In particular, $V$ commutes with $C$ because $C$ is a polynomial in $V$. On each common eigendirection, $v>0$ solves $v^{2}+p^{2}\Delta v-c=0$, which has exactly one positive root for $c>0$. Functional calculus gives~\eqref{eq:matrix-gauge-optimum}, including repeated eigenvalues.

For completeness, a global optimum exists. If $s_{B},s_{A}>0$ are lower spectral bounds for the two moment matrices and $\lambda_{i}$ are eigenvalues of $\Gamma$, the nonconstant objective is bounded below by
\[
\sum_{i=1}^{r}\left[\frac{s_{B}\lambda_{i}+s_{A}/\lambda_{i}}{2p^{2}}+\frac{\Delta}{2}\log\lambda_{i}\right].
\]
Each summand is bounded below and diverges as its argument tends to zero or infinity. Sublevel sets are therefore compact inside the positive-definite cone. Every minimizer is stationary, and the stationary solution is unique. Any factor $R$ with $RR^{\top}=\Gamma_{*}$ attains it, explaining the remaining orthogonal freedom.

Finally, $(BR)(R^{-1}A)=BA$ for every draw, not merely in expectation. A fixed base matrix and LoRA scaling preserve this equality. The full network prediction is consequently unchanged when every transformed adapter is used with its paired factors. Replacing the transformed Gaussian by a new isotropic distribution would discard this guarantee.

\subsection{Scalar special case}
\begin{proposition}[Scalar gauge envelope]
\label{prop:gauge-envelope}
Let $U_{B}=\E\norm {B}_{F}^{2}$, $U_{A}=\E\norm {A}_{F}^{2}$, $d_{B}=mr$, and $d_{A}=rn$. For $(B,A)\mapsto(cB,A/c)$ with $c>0$, the unique optimum is
\begin{equation}
c_{*}^{2}=\frac{-p^{2}(d_{A}-d_{B})+\sqrt{p^{4}(d_{A}-d_{B})^{2}+4U_{A}U_{B}}}{2U_{B}}.
\label{eq:gauge-optimum}
\end{equation}
\end{proposition}
Substitute $\Gamma=c^{2}I$ into~\eqref{eq:matrix-gauge-kl}. Its nonconstant terms are $(c^{2}U_{B}+c^{-2}U_{A})/(2p^{2})+(d_{A}-d_{B})\log c$. Differentiation and $x=c^{2}$ give $U_{B}x^{2}+p^{2}(d_{A}-d_{B})x-U_{A}=0$. The only positive root is~\eqref{eq:gauge-optimum}. The matrix family contains this scalar family, so its optimum cannot be worse. At $r=1$ they coincide.

\begin{algorithm}[t]
\caption{Behavior-preserving matrix balancing}
\label{alg:matrix-gauge}
\begin{algorithmic}[1]
\Require posterior moments $S_{B},S_{A}\succ0$, factor dimensions $m,n$, fixed prior width $p$
\State $C\gets S_{B}^{1/2}S_{A}S_{B}^{1/2}$; diagonalize $C=V\operatorname{diag}(c_{i})V^{\top}$
\State $d_{0}\gets p^{2}(n-m)$
\For{each eigenvalue $c_{i}$}
\If{$d_{0}\ge0$}
\State $v_{i}\gets 2c_{i}/(\sqrt{d_{0}^{2}+4c_{i}}+d_{0})$
\Else
\State $v_{i}\gets (\sqrt{d_{0}^{2}+4c_{i}}-d_{0})/2$
\EndIf
\EndFor
\State $\Gamma_{*}\gets S_{B}^{-1/2}V\operatorname{diag}(v_{i})V^{\top} S_{B}^{-1/2}$; $R\gets \Gamma_{*}^{1/2}$
\State transform each complete draw to $(BR,R^{-1}A)$ and evaluate KL by~\eqref{eq:matrix-gauge-kl}
\end{algorithmic}
\end{algorithm}
The matrix operations are on $r\times r$ moments. Forming moments has costs depending on the factor representation; the spectral step costs $O(r^{3})$. This is an arithmetic complexity statement, not a measured language-model latency gain.

\section{Noncentral gauge geometry and fixed-predictor diagnostics}
\label{app:noncentral-proof}
\subsection{The fixed noncentral Gaussian identity}
Let $P_{0}=\mathcal {N}((B_{0},A_{0}),p^{2}I_{d})$, where $(B_{0},A_{0})$ and $p$ are fixed independently of the certification sample. Its negative log density at $(B',A')$ is
\[
 \frac {d}{2}\log(2\pi p^{2})+\frac{\norm{B'-B_{0}}_{F}^{2}+\norm{A'-A_{0}}_{F}^{2}}{2p^{2}}.
\]
The posterior entropy after the linear change of variables is
$h(Q_{R})=h(Q)+(m-n)\log|\det R|$. Substitution of $B'=BR$, $A'=R^{-1}A$ proves~\eqref{eq:noncentral-gauge}. This calculation allows correlation between the two Gaussian factors and transforms the full covariance, not just its marginal variances.

For the polar parameterization $R=\Gamma^{1/2}O$, $R^{-1}=O^{\top} \Gamma^{-1/2}$. The noncentral cross terms become
\begin{align*}
 \langle M_{B}R,B_{0}\rangle_{F}
 &=\operatorname{tr}\!\left(O^{\top} \Gamma^{1/2}M_{B}^{\top} B_{0}\right),\\
 \langle R^{-1}M_{A},A_{0}\rangle_{F}
 &=\operatorname{tr}\!\left(O^{\top} \Gamma^{-1/2}M_{A}A_{0}^{\top}\right).
\end{align*}
The remaining quadratic terms are $\operatorname{tr}(S_{B}\Gamma)$ and $\operatorname{tr}(S_{A}\Gamma^{-1})$. Thus the orientation enters only through $-2\operatorname{tr}(O^{\top} F(\Gamma))$. If $F=L\Lambda V^{\top}$, orthogonality gives
\[
 \operatorname{tr}(O^{\top} F)=\operatorname{tr}\!\left(V^{\top} O^{\top} L\Lambda\right)
 \le\sum_{i}\Lambda_{ii}=\norm {F}_{*}.
\]
Taking $O=LV^{\top}$ attains equality, including when $F$ is rank deficient. This proves Proposition~\ref{prop:noncentral-polar}.

To prove existence over $\Gamma$, define the centered second moments
$G_{B}=\E[(B-M_{B})^{\top}(B-M_{B})]\succ0$ and
$G_{A}=\E[(A-M_{A})(A-M_{A})^{\top}]\succ0$. Positive definiteness follows from nonsingularity of the joint Gaussian. Variance decomposition gives
\[
 \E\norm{BR-B_{0}}_{F}^{2}\ge\operatorname{tr}(G_{B}\Gamma),\qquad
 \E\norm{R^{-1}A-A_{0}}_{F}^{2}\ge\operatorname{tr}(G_{A}\Gamma^{-1}).
\]
For lower eigenvalue bounds $g_{B},g_{A}>0$, the objective is therefore bounded below by a constant plus
\[
 \sum_{i}\left[\frac{g_{B}\lambda_{i}+g_{A}/\lambda_{i}}{2p^{2}}
 +\frac{n-m}{2}\log\lambda_{i}\right],
\]
where $\lambda_{i}$ are the eigenvalues of $\Gamma$. Each summand diverges at zero and infinity. The reduced objective is continuous, so it attains a minimum on a compact sublevel set. This is an existence proof, not a convexity or uniqueness argument. At $B_{0}=A_{0}=0$, $F(\Gamma)=0$ and the zero-centered theorem is recovered exactly.

\subsection{Numerical use without assuming the zero-centered optimizer}
The identity transformation is always a valid candidate. For a fixed $\Gamma$, the polar step minimizes over the complete orthogonal group, including both determinant signs. One may optimize the remaining objective over a positive-definite parameterization, compare all attained values with the identity, and retain the smallest actual Gaussian KL. Rank deficiency of $F(\Gamma)$ can make the nuclear-norm term nondifferentiable; numerical stationarity is not a global-optimality certificate. The exact zero-centered optimizer is a possible initialization, not a substitute for the noncentral objective.

For a direct matrix-variable check, let $L_{B}=M_{B}^{\top} B_{0}$, $L_{A}=M_{A}A_{0}^{\top}$, and $\Delta=n-m$. At nonsingular $R$, differentiation of~\eqref{eq:noncentral-gauge} gives
\begin{align}
 \nabla_{R}\KL(Q_{R}\|P_{0})
 ={}&p^{-2}\!\left[S_{B}R-L_{B}-R^{-\top}R^{-1}S_{A}R^{-\top}
 +R^{-\top}L_{A}^{\top} R^{-\top}\right]
 +\Delta R^{-\top}.
 \label{eq:noncentral-gradient}
\end{align}
This gradient evaluates the original smooth objective on $GL(r)$, whereas the reduced objective has already optimized out the orthogonal factor. For each retained $R$, paired sampling must use $(BR,R^{-1}A)$ exactly. Replacing this pushforward by a freshly isotropic Gaussian changes the stochastic predictor.

For any invertible measurable $T$ applied to both distributions, the change-of-variables Jacobians cancel in their density ratio:
\[
 \KL(T_{\#}Q\|T_{\#}P_{0})=\KL(Q\|P_{0}).
\]
The improvement studied here instead keeps $P_{0}$ fixed and changes the parameter distribution within a prediction-equivalent orbit. A zero initial $B_{0}$ does not restore the zero-centered theorem when $A_{0}\ne0$: the surviving term $\Gamma^{-1/2}M_{A}A_{0}^{\top}$ still determines a preferred orientation.

\subsection{A valid initialization-centered comparison}
A matched comparison must save the actual initialization of every trained block, including the classification head, before learning from $S$. For a posterior with block widths $q_{b}$, the two candidate prior centers use the same posterior and therefore exactly the same predictions. Their Gaussian KLs differ only in the squared displacements when prior covariances are identical. A posterior selected for prediction stability is a separate choice: changing $q_{b}$ changes both its loss and covariance cost, so the two choices cannot be combined into an attributed gain without the matched comparisons.

If both origin and initialization centers are selected together with the nine rank--regularization configurations, a uniform declared mixture has 18 components, not nine. The width search remains posterior-side when the prior is fixed. Final certification draws must be independent of that search, or a simultaneous bound must cover all inspected candidates. Hoeffding and Chernoff comparisons use the same draw-level losses; selecting between them requires covering both unless the inequality was fixed independently of those draws. All these comparisons can use Theorem~\ref{thm:pac-bayes-kl}, but none supplies an initialization-centered empirical result in the absence of the corresponding checkpoints and draws.

\subsection{Reproducing the accounting floors from recorded summaries}
\label{app:accounting-floor}
At $K_{\rm MC}=9$ and $\eta=.025$, let $c=\log360$. For $M=8$, clipping $\widetilde {r}+\sqrt{c/16}$ to one implies
$\widetilde {r}\ge 0.3934676414$. Hence
\[
 U_{\rm kl}\ge\klinverse(1-\sqrt{c/16},c/8)
 =0.8941321961\ldots.
\]
For $M=64$, the conservative lower endpoint of the reported three-decimal interval is $0.6715$. Since it is below one, clipping is inactive at that minimum. Therefore every recorded mean satisfies
\[
 \widetilde {r}\ge0.6715-\sqrt{c/128}=0.4570584281\ldots,
\]
which yields $U_{\rm kl}\ge0.6662649085\ldots$. These inequalities use the stated rounded summaries, not unobserved per-configuration means.

The function
$[\log9+\log(2\sqrt {N}/.025)]/N$ decreases for $N\ge2$. Thus $N\le1024$ permits the common favorable confidence cost
\[
 c_{\rm PB,0}=\frac{\log9+\log(64/.025)}{1024}.
\]
Applying $\klinverse(\cdot,c_{\rm PB,0})$ gives lower bounds
$0.9320526640\ldots$ and $0.7298471427\ldots$, respectively. Table~\ref{tab:accounting-floor} rounds downward. Both exceed $0.5$, even though KL is optimistically set to zero. Enlarging the confidence family increases these floors. Increasing $M$ with newly observed losses, changing posterior widths, or changing the empirical-risk estimator is outside this fixed-summary calculation.

\section{Language-model experimental details}
\label{app:extended-experiments}
\begin{table}[t]
\centering\small
\caption{Paired selector experiment. Requested training-pool sizes are capped by the available split. Confidence guarantees hold separately for each selection problem.}
\label{tab:protocol}
\begin{tabular}{p{.24\linewidth}p{.68\linewidth}}
\toprule
Component & Setting\\
\midrule
Backbone & RoBERTa-base and its tokenizer; fitted sequence-classification head\\
Training pools & Requested sizes $256,512,1024$; three seeds $0,1,2$; seven tasks\\
Sequence processing & Maximum 128 tokens; dynamic padding\\
Optimization & AdamW, base learning rate $5\times10^{-5}$, head multiplier 10; three epochs with a floor of 400 optimizer steps; batch size 16\\
Adapters & Query/value matrices; ranks $4,16,64$; scaling $\alpha=2r$; dropout zero; no adapter bias\\
Regularization & LoRA decay $10^{-3},10^{-2},10^{-1}$; other trained blocks have zero decay\\
Prior & Origin-centered isotropic Gaussian with $p=1/\sqrt{\lambda}$; nine components with weight $1/9$\\
Posterior & Gaussian on all trained coordinates; one width selected from seven candidates before final independent draws\\
Confidence & $\delta_{\rm PB}=.025$, $\eta=.025$, $K_{\rm MC}=9$ for a frozen final posterior per configuration\\
Evaluation & Public development examples, capped at 800; MRPC 408, RTE 277, WNLI 71\\
Hardware & One NVIDIA A10G\\
Inference & Paired task means; all $2^{7}$ sign flips; 10,000 task-bootstrap draws\\
\bottomrule
\end{tabular}
\end{table}
The center-risk proxy is explicitly
\[
\widehat {R}_{S}(\delta_{\mu_{k}})+
\sqrt{\frac{\KL_{\rm LoRA}+\log(2\sqrt {N}/.05)}{2(N-1)}}.
\]
It omits the fitted head, hierarchical penalty, and Monte Carlo upper bound. The literal certificate uses~\eqref{eq:audited-selector} with the actual sample size. The historical width search has seven candidates, but their numerical values and whether they were absolute or prior-relative are not recorded in the available experimental specification. This prevents exact replication of that historical search. The explicitly listed nine-width frontier and controlled-study grids below are different experiments and do not reconstruct it. The certified dimension is obtained from all trained blocks rather than an adapter-only name filter.

Validation and five-fold cross-validation select configurations without access to evaluation labels. The retrained validation baseline uses the entire training pool only after selection. The unretrained validation model reaches $0.695$ mean accuracy, compared with $0.705$ for the proxy, but this comparison gives the proxy more fitting data. It is not the primary comparison. A test-label oracle has center accuracy $0.7339$ and stochastic accuracy $0.5145$; it is an upper reference, not an admissible selector. Complexity-only selection coincides with the smallest-rank rule in this audit.

\begin{table}[t]
\centering\small
\caption{Secondary paired comparisons. Intervals are descriptive 95\% task-cluster bootstrap intervals. Cross-validation rows use only their matched 21-problem grid. Native metrics re-score frozen accuracy-selected configurations; selection was not rerun with a different metric.}
\label{tab:selector-audit}
\begin{tabular}{lrl}
\toprule
Selector on matched 21 problems & Accuracy & Effect relative to validation\\
\midrule
Five-fold cross-validation & .7180 & $+.0062\;[-.0059,.0209]$\\
Center-risk proxy & .7166 & $+.0047\;[-.0046,.0132]$\\
Proxy minus cross-validation & -- & $-.0014\;[-.0225,.0142]$\\
\bottomrule
\end{tabular}
\medskip
\begin{tabular}{llrrr}
\toprule
Task & Metric & Proxy & Training risk & Validation\\
\midrule
CoLA & Matthews correlation & .416 & .412 & .379\\
MRPC & F1 & .847 & .847 & .846\\
QQP & F1 & .654 & .653 & .664\\
\bottomrule
\end{tabular}
\label{tab:task-native-metrics}
\end{table}

\subsection{Secondary selector comparison}

RoBERTa-base \citep{liu2019roberta} is fine-tuned on seven GLUE tasks \citep{wang2019glue}: CoLA, MRPC, QNLI, QQP, RTE, SST-2, and WNLI. A selection problem combines one task, a requested training pool in $\{256,512,1024\}$, and a seed in $\{0,1,2\}$. Each problem contains the nine combinations of rank $\{4,16,64\}$ and LoRA weight decay $\{10^{-3},10^{-2},10^{-1}\}$, for 63 problems and 567 configurations. Query and value adapters and the classification head are trained. Evaluation uses the public development split, capped at 800 examples; actual smaller sets contain 408 MRPC, 277 RTE, and 71 WNLI examples. WNLI has only 635 training examples, so requested and actual pool sizes are distinguished.

The center-risk proxy adds a Gaussian-KL square-root penalty to deterministic training error, with only LoRA factors charged. Its prior is $\mathcal {N}(0,\lambda^{-1}I)$ and its risk input is not posterior-averaged. The valid certificate instead charges every trained coordinate, uses a uniform nine-configuration prior, and estimates stochastic risk with independent final draws. The total per-problem failure probability is $0.05$, split equally between PAC-Bayes and Monte Carlo. Posterior widths are selected before final certification; Table~\ref{tab:protocol} lists the protocol, and the limitations of its recorded settings follow the table.

Validation selects on a $25\%$ training-pool holdout and then retrains the chosen configuration on the complete pool. This is the appropriate comparison with full-pool-trained checkpoints, not the weaker unretrained validation model. Effects are paired within problems and reduced to seven task means. Exact sign flips use task-level effects and require joint sign symmetry under the null; task-cluster bootstrap intervals are descriptive because the benchmark tasks are not sampled from a defined population of tasks.

\begin{table}[t]
\centering\small
\setlength{\tabcolsep}{3.1pt}
\caption{Selection on 63 problems. Accuracy is evaluated at the selected deterministic center and under its posterior. Effects are relative to validation with retraining, with descriptive 95\% task-cluster bootstrap intervals. Wins count tasks, not individual configurations.}
\label{tab:main-results}
\begin{tabular}{lrrlr}
\toprule
Selector & Center & Stochastic & Effect [95\% interval] & Wins\\
\midrule
Center-risk Gaussian-KL proxy & .7053 & .5107 & $-.0035\;[-.0209,.0070]$ & 5/7\\
Training risk & .7066 & .5186 & $-.0023\;[-.0206,.0086]$ & 5/7\\
Validation, retrained & .7088 & .5152 & $0$ & --\\
BIC & .6883 & .5235 & $-.0205\;[-.0373,-.0002]$ & 1/7\\
Smallest rank & .6877 & .5229 & $-.0211\;[-.0375,-.0017]$ & 1/7\\
\bottomrule
\end{tabular}
\end{table}

\begin{figure}[ht]
\centering
\includegraphics[width=\linewidth]{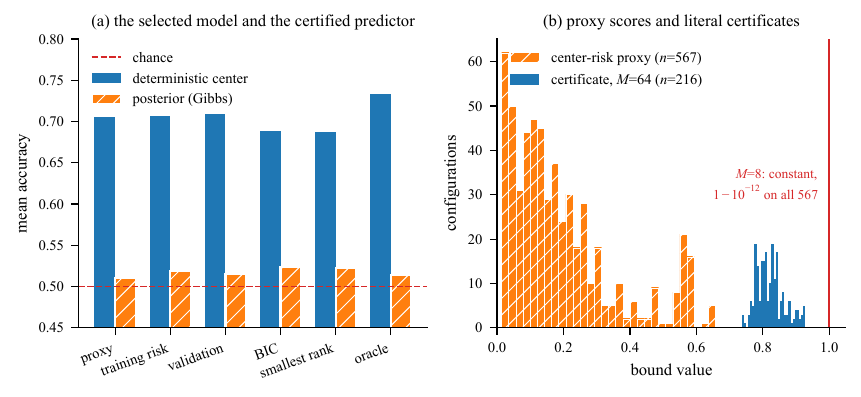}
\caption{Selected models versus the certified predictor on the 63 small-pool problems. (a) Mean evaluation accuracy of each selector's chosen configuration at its deterministic center and under its Gaussian posterior; the oracle uses test labels and is not an admissible selector. (b) Center-risk proxy scores on all 567 configurations and literal 64-draw certificates on the matched 216; every 8-draw certificate equals one.}
\label{fig:selector-certificate}
\end{figure}

The proxy reaches $0.7053$ mean accuracy, below training risk ($0.7066$) and retrained validation ($0.7088$). Its task-level sign-flip $p$-value against validation is $0.9844$; this does not establish equivalence. The more important difference is between predictors: the selected centers are substantially more accurate than their associated stochastic models (Table~\ref{tab:main-results}). A detailed MRPC cell has center and posterior training errors $0.0703$ and $0.6040$, respectively. On evaluation examples their errors are $0.1912$ and $0.6379$, with $54.2\%$ disagreement. The sampled models collapse to constant-class predictions. A small Gaussian KL therefore coexists with an unusable stochastic predictor.
Figure~\ref{fig:selector-certificate} shows the same gap for every selector, together with the distributions of proxy scores and literal certificates.

\subsection{Historical draw-budget comparison}

At $M=8$ and $K_{\rm MC}=9$, the Hoeffding half-width is $0.6065$. Added to the observed posterior errors, it clips every empirical-risk upper bound to one. All 567 certificates consequently equal one mathematically; the stored inversion clips to $1-10^{-12}$. Every problem has a nine-way tie, and enumeration selects the smallest rank in 63/63 problems. The complete KL is only $2.34$--$45.91$ nats, with mean $\KL/N=0.0289$. Monte Carlo uncertainty, rather than complexity, causes this saturation.

\begin{table}[t]
\centering\small
\caption{The original eight-draw certificate and a matched 64-draw rerun. The latter covers 216 configurations in 24 problems. Selected accuracies in the last row are compared only within this common subset; stochastic evaluation predictions were not recorded for the rerun.}
\label{tab:paper-cert}
\begin{tabular}{lrrr}
\toprule
Quantity & Center-risk proxy & Certificate, $M=8$ & Certificate, $M=64$\\
\midrule
Charged blocks & LoRA only & All trained & All trained\\
Mean KL (nats) & 4.29 & 12.49 & 12.67\\
Risk input & Center & Upper bound $1$ & Upper bound $.672$--$.761$\\
Score or certificate & Mean $.1935$ & Constant $1$ & $.736$--$.928$\\
Selected accuracy, matched 24 & .7634 & .7257 & .7388\\
\bottomrule
\end{tabular}
\end{table}

With $M=64$, the half-width falls to $0.2144$. All 216 certificates in the matched rerun are distinct, with range $0.736$--$0.928$. Their selected center accuracy is $0.7388$, below the proxy ($0.7634$), training risk ($0.7655$), and retrained validation ($0.7582$) on those same 24 problems. Removing a tie does not make the certificate a better selector. Conversely, near-one values alone do not prove poor ordering: selected-model accuracy must be evaluated explicitly.

\subsection{What tighter accounting can change on the recorded LoRA grid}

The small KL in Table~\ref{tab:paper-cert} makes the complexity-free floor directly relevant. At $M=8$, Hoeffding saturation implies
$\widetilde {r}\ge1-\sqrt{\log(360)/16}$. Even the more favorable single-inequality Chernoff charge $K_{\rm MC}=9$ then gives $U_{\rm kl}\ge0.8941$. For the 64-draw grid, the smallest reported Hoeffding upper bound is $0.672$. Taking its three-decimal rounding conservatively as $0.6715$ gives
\begin{equation}
 \widetilde {r}\ge0.6715-\sqrt{\frac{\log360}{128}},\qquad
 U_{\rm kl}\ge0.6662.
 \label{eq:recorded-risk-floor}
\end{equation}
Because every training pool has $N\le1024$, Equation~\eqref{eq:zero-kl-floor} strengthens these to the conservative certificate floors in Table~\ref{tab:accounting-floor}. Thus even a hypothetical zero-KL representation, combined with Chernoff instead of Hoeffding, cannot meet $\tau=0.5$ at either recorded budget. This conclusion holds before estimating whether real trained factors benefit from matrix balancing.

\begin{table}[t]
\centering\small
\caption{Arithmetic lower bounds for the recorded LoRA grids. Each entry is a conservative lower bound, not a rerun mean or a certificate issued after setting KL to zero. The calculation grants the favorable $N=1024$, uses $w_{k}=1/9$ and $\delta_{\rm PB}=\eta=.025$, and rounds downward.}
\label{tab:accounting-floor}
\begin{tabular}{rrrr}
\toprule
Draws & Configurations & Chernoff risk UCB floor & Zero-KL certificate floor\\
\midrule
8 & 567 & $0.8941$ & $0.9320$\\
64 & 216 & $0.6662$ & $0.7298$\\
\bottomrule
\end{tabular}
\end{table}

The floors compare accounting on the same realized draw losses; they are not lower confidence bounds on population risk. More draws or a different posterior can change them. Inspecting both inequalities with a simultaneous guarantee requires an additional Monte Carlo allocation, which can only increase these optimistic floors. In particular, an initialization-centered prior cannot rescue these fixed losses through a smaller KL alone, although initialization-centered, narrower posteriors remain a different and unmeasured comparison.

\begin{figure}[ht]
\centering
\includegraphics[width=\linewidth]{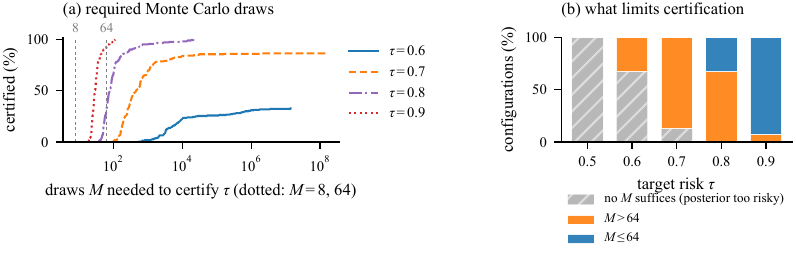}
\caption{Monte Carlo draws needed on the 216 recorded 64-draw configurations, with each configuration's KL and observed mean held fixed and the Hoeffding bound at $K_{\rm MC}=9$. (a) Cumulative share certifiable at target $\tau$ as a function of the required draws $M$; curves stop below 100\% when no finite $M$ suffices. (b) Binding constraint per target: no finite $M$ suffices (observed mean not below $u_{\tau}$, or no $u_{\tau}$ exists), more than 64 draws, or at most 64 draws.}
\label{fig:draw-budget}
\end{figure}

Figure~\ref{fig:draw-budget} inverts the Hoeffding budget~\eqref{eq:mc-budget} on the 216 recorded 64-draw configurations, holding each configuration's complete KL and observed mean fixed. At $\tau=0.5$ no finite number of draws suffices for any configuration, because the observed mean is not below the allowed risk $u_{\tau}$ or no such risk exists. At $\tau=0.6$ this holds for 145 configurations and at $\tau=0.7$ for 28, with median requirements of 6137 and 343 draws among the rest. Only at $\tau=0.8$ and $0.9$ do 64 draws suffice for 71 and 201 configurations. This is fixed-summary arithmetic, not a sequentially valid stopping rule.

\subsection{Family exclusion and a behavior-preserving KL correction}

For each of the 567 configurations, we evaluate Corollary~\ref{cor:isotropic-width} with the origin-centered prior, $q_{\max}=1$, and the complete dimension. The exclusion counts at risk targets $\tau=0.5,0.4,0.3,0.2,0.1$ are, respectively, $243,261,429,511,567$. These exclusions cover all widths in $(0,1]$, not just sampled grid points. At the empirical training-pool majority-error reference, $297/567$ configurations are excluded; that reference is not itself a certified population risk. A nine-width risk sweep has a simultaneous half-width of $0.913$, giving zero lower bounds throughout and no sharper exclusions.

\begin{figure}[ht]
\centering
\includegraphics[width=.62\linewidth]{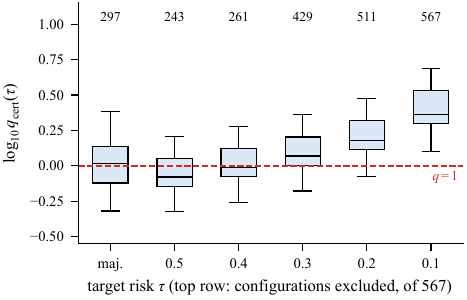}
\caption{Prior-scale-free width threshold $q_{\rm cert}(\tau)$ over the 567 configurations (origin-centered prior, complete dimension). Boxes show quartiles and whiskers $1.5$ interquartile ranges; values above the dashed line exclude every $q\in(0,1]$. The top row gives the excluded count; ``maj.'' uses each configuration's training-pool majority error as its target.}
\label{fig:qcert-dist}
\end{figure}

Figure~\ref{fig:qcert-dist} shows the full distribution of $q_{\rm cert}(\tau)$ behind these counts. The median threshold moves above $q=1$ as the target tightens, so the exclusions at small $\tau$ are not driven by a few extreme configurations.

\subsection{Task dependence and the WNLI sensitivity analysis}
The exact two-sided task sign-flip $p$-values are $0.9844$ for the proxy and training-risk selector, and $0.1094$ for BIC and the smallest rank. Equal absolute statistics are included in the tail. These tests rely on joint sign symmetry and do not transform related benchmark tasks into an independent random sample of NLP problems.

Excluding WNLI post hoc reverses the proxy's mean effect to $+0.0048$ with interval $[0.0016,0.0080]$ and $p=0.0625$. Training risk then has effect $+0.0065$, interval $[0.0030,0.0095]$, and the same $p$-value. WNLI has only 71 evaluation examples and every evaluated selector is below its majority baseline. The exclusion is a sensitivity analysis, not a prospective claim of superiority on a newly defined task population.

\begin{figure}[ht]
\centering
\includegraphics[width=\linewidth]{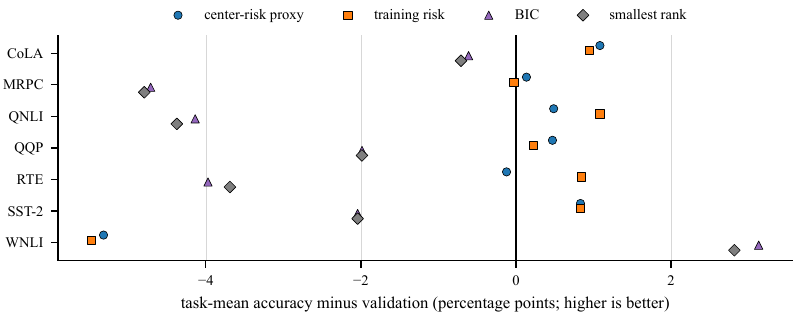}
\caption{Task-level selector effects relative to retrained validation, in percentage points of task-mean accuracy over the nine problems of each task (higher is better). WNLI, with 71 evaluation examples, drives the negative mean effect of the proxy and training-risk selectors.}
\label{fig:task-effects}
\end{figure}

\subsection{Posterior and confidence diagnostics}
\label{app:selector-results}\label{app:isotropic-screen}
On the MRPC cell with $N=512$, rank 16, and decay $10^{-2}$, posterior draws attain the two constant-class evaluation risks $0.3162$ and $0.6838$. The $M=16$ posterior-risk upper bound is $0.9435$; replacing the center-risk input by this upper bound moves the inverse-KL expression from $0.1251$ to $0.9757$. The latter is mathematically below one but does not establish a useful classifier.

The $M=64$ rerun uses CoLA, MRPC, RTE, and SST-2. A more conservative variant uses $K_{\rm MC}=63$ and also charges $\log(63/9)$ extra nats in the mixture penalty. It yields $0.772$--$0.946$ and the same selected configuration as $K_{\rm MC}=9$ on 23 of 24 problems. This variant is not needed when width selection precedes a fresh final stream, but shows the consequence of a larger declared family.

The width screen uses $D=\norm{\mu}^{2}$ and $a_{k}=\log9+\log(2\sqrt {N}/.025)$. A consistency check between the analytic endpoint at $q=1$ and nine logarithmic widths in $[10^{-4},1]$ yields no disagreements. The mean measured posterior risk at $q=1$ is $0.50$, but the simultaneous half-width $0.913$ makes all risk lower bounds zero. These risk measurements therefore do not contribute to the exclusion counts. Initialization-centered priors and widths above one remain outside the reported exclusion.

\section{Controlled numerical studies}
\label{app:controlled}
\subsection{Gaussian-factor classification}
For each of 20 seeds, sample $N=2048$ independent $x_{i}\sim\mathcal {N}(0,I_{2})$ and labels $y_{i}=\operatorname{sign}(x_{i1})$. Define $\bar {a}=N^{-1}\sum_{i} y_{i}x_{i}$ and $\mu_{A}=2\bar {a}/\norm{\bar {a}}$. The posterior has independent factors $B\sim\mathcal {N}(1,q^{2})$ and $A\sim\mathcal {N}(\mu_{A},q^{2}I_{2})$, and predicts $\operatorname{sign}(BAx)$. The fixed prior is $\mathcal {N}(0,I_{3})$. Its exact KL is
\[
\KL(Q_{q}\|P)=\tfrac{1}{2}\left[1+\norm{\mu_{A}}^{2}+3q^{2}-3-3\log q^{2}\right].
\]
The center is fitted on $S$, while the prior is independent of it. A uniform configuration label costs $\log9$ conservatively; all widths share the same prior, so this penalty is not mathematically necessary for posterior-width selection.

The widths are $\{.05,.1,.2,.4,.8,1.2,2,4,8\}$. Each receives 1024 independent draws, with fixed nested prefixes of sizes $8,64,256,1024$. We allocate $\eta=.025$ across nine widths, four budgets, and two inequalities, so both reported Monte Carlo layers use $\log(72/.025)$. The PAC-Bayes allocation is $\delta_{\rm PB}=.025$. All certificate arithmetic uses~\eqref{eq:audited-selector} and the same sampled predictions within a comparison.

The empirical stochastic risk has an analytic reference. Set $b=\Phi(-1/q)$ and $a_{i}=\Phi(-y_{i}\mu_{A}^{\top} x_{i}/(q\norm{x_{i}}))$. Independence of $B$ and $A$ gives
\begin{equation}
\widehat {R}_{S}(Q_{q})=b+(1-2b)\frac{1}{N}\sum_{i} a_{i}.
\label{eq:exact-control-risk}
\end{equation}
This formula checks Monte Carlo coverage without treating a longer Monte Carlo stream as ground truth. Across the $720$ width--budget--seed rows, neither upper-bound implementation undercovers this reference. This is an implementation check, not empirical proof of its nominal coverage. The sample SD in Table~\ref{tab:mc-control} includes dataset and posterior-sampling variation through the independent seeds; it is not a standard error.

\subsection{Matrix-gauge construction and independent optimization}
We use $(m,n)\in\{(12,8),(8,12),(12,12)\}$, ranks $r\in\{1,2,4,8\}$, and seeds $0$--$19$. A random orthogonal matrix $O$ rotates a deformation with eigenvalues $\exp(\operatorname{linspace}(-2,2,r))$; for rank one the eigenvalue is $e^{2}$. Let this deformation be $G$. Independent Gaussian arrays generate $M_{B}=(G_{B}/\sqrt {m})G$ and $M_{A}=G^{-1}(G_{A}/\sqrt {n})$. The posterior uses independent isotropic factor noises $s_{B}=.15,s_{A}=.10$ and the prior width is $p=1$. Thus
\[
S_{B}=M_{B}^{\top} M_{B}+ms_{B}^{2}I,\qquad S_{A}=M_{A}M_{A}^{\top}+ns_{A}^{2}I.
\]
The deformation deliberately makes scalar and matrix balancing distinguishable; it is not a claim about the distribution of trained language-model factors.

For every case, we compute raw, scalar, and matrix-optimal KL. An independent BFGS solve parameterizes $\Gamma=LL^{\top}$ with exponential diagonal entries in $L$, starts from the identity, and minimizes the same objective with an analytic gradient, tolerance $10^{-8}$, and at most 1000 iterations. We report the objective difference, not a blanket optimizer-convergence assertion. The maximum relative difference, normalized by $1+|\KL_{*}|$, is $9.91\times10^{-16}$. The matrix stationarity residual is normalized by $1+\norm{S_{A}}_{F}$.

For each case, 128 independent factor draws are paired before and after the exact pushforward. Product discrepancy is $\norm{B'A'-BA}_{F}/(1+\norm{BA}_{F})$. Its maximum is $2.71\times10^{-15}$. The inclusion ordering $\KL_{\rm matrix}\le\KL_{\rm scalar}\le\KL_{\rm raw}$ is checked in every row. Seed-level outputs and complete summaries for all three matrix shapes accompany the code.

\section{Split-prior numerical evaluation scenarios}
\label{app:large-task-protocol}
Displacements, stochastic losses, matrix-KL reductions, and decoded errors are assumed inputs; the raw Gaussian KLs, confidence bounds, width thresholds, certification margins, and evaluation counts are calculated from them.

\subsection{Independent prior and posterior specification}
The split-prior design uses RoBERTa-base with rank-four query and value adapters, a prior-data partition of 33,000 units, and nominal certification-sample sizes $N\in\{1000,4000,16000,33000\}$ for SST-2 and $N\in\{1000,4000,16000,70000\}$ for QNLI. These are numerical design sizes, not verified independent-unit counts. The classification head is fitted only on the prior partition and is frozen before certification data are accessed. Query and value factors across the twelve width-768 layers contain 147,456 certified coordinates. Source-level independence and available sample counts remain conditions of the application; row disjointness alone does not establish independence.

The training specification uses seeds $0,1,2$, three prior-training epochs, maximum sequence length 128, batch size 16, LoRA scale $\alpha=2r$, zero adapter dropout, and AdamW. The prior-stage factor and head learning rates are $5\times10^{-5}$ and $5\times10^{-4}$, with factor weight decay $10^{-2}$. Each continued checkpoint starts from its saved prior center and takes 400 factor-only steps at learning rate $2\times10^{-6}$ with zero weight decay. Both prior factors can be nonzero after this training. The numerical tables instantiate single assumed profiles, not averages over executed seed replicates.

Prior widths belong to $\{.005,.01,.02,.04\}$ with mixture weights $1/4$. The posterior-width grid is $\{.005,.01,.015,.02,.025,.03,.04\}$. A pilot searches the Cartesian product of the factor-group widths using 32 draws on at most 1024 certification examples. The final family retains the seven equal-width profiles, twelve profiles varying one group around $.02$, and at most four additional pilot choices, for at most 23 distinct profiles. All final profiles are frozen before independent final draws.

The confidence allocation for each task--seed study is $.02$ for Gaussian PAC-Bayes, $.02$ for Monte Carlo, $.005$ for compression, and $.005$ for the fixed-label reference. Division over the four sample sizes gives $\delta_{\rm PB}=\eta=.005$ and $\delta_{\rm code}=\delta_{\rm base}=.00125$ per size. The Monte Carlo family covers 23 profiles, two fixed budgets, and both inequalities, giving $K_{\rm MC}=92$. Balancing adds no draw-level multiplicity when it preserves the complete paired stochastic predictor. The principal tables use $M=1024$. Guarantees across additional task--seed studies require their own allocation.

\subsection{Sample-size frontier}
Table~\ref{tab:large-task-record} instantiates the equal-width case $p=q_A=q_B=.02$. Its raw KL is exactly $D/(2p^{2})$; no covariance-mismatch term is omitted. The Chernoff upper bound is computed from the assumed draw mean with the full final-family charge. The matrix-KL entries are stipulated attainable values, not values derived from the scalar displacement alone. Their use in the certificate is conditional on a valid noncentral pushforward attaining them.
\begin{table}[t]
\centering\small
\setlength{\tabcolsep}{3.6pt}
\caption{Split-prior sample-size frontier. $\widetilde r$ and $D$ are assumed, $K_R$ is an assumed attained matrix KL, and $U_{\rm kl}$ and $C_R$ are calculated. KL is in nats; risks are proportions. The same posterior losses apply before and after a valid pushforward.}
\label{tab:large-task-record}
\begin{tabular}{lrrrrrrr}
\toprule
Task & $N$ & $D$ & $\widetilde r$ & $U_{\rm kl}$ & $K_I$ & $K_R$ & $C_R$\\
\midrule
SST-2 & 1{,}000 & 0.42 & 0.059 & 0.0973 & 525.00 & 490.35 & 0.5719\\
SST-2 & 4{,}000 & 0.46 & 0.066 & 0.1059 & 575.00 & 547.40 & 0.3321\\
SST-2 & 16{,}000 & 0.50 & 0.071 & 0.1121 & 625.00 & 601.25 & 0.2182\\
SST-2 & 33{,}000 & 0.54 & 0.076 & 0.1181 & 675.00 & 652.73 & 0.1928\\
QNLI & 1{,}000 & 0.62 & 0.081 & 0.1241 & 775.00 & 713.77 & 0.7003\\
QNLI & 4{,}000 & 0.70 & 0.089 & 0.1336 & 875.00 & 821.62 & 0.4314\\
QNLI & 16{,}000 & 0.82 & 0.098 & 0.1442 & 1025.00 & 976.82 & 0.2939\\
QNLI & 70{,}000 & 0.90 & 0.107 & 0.1547 & 1125.00 & 1081.12 & 0.2255\\
\bottomrule
\end{tabular}
\end{table}

\begin{figure}[ht]
\centering
\includegraphics[width=\linewidth]{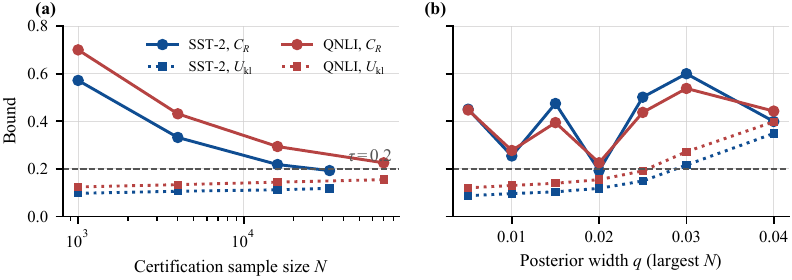}
\caption{Split-prior scenarios (assumed inputs; values are formula-derived, not measured). (a) Balanced certificate $C_R$ and Chernoff risk bound $U_{\rm kl}$ versus certification sample size at $p=q=.02$. (b) The same quantities versus posterior width at the largest $N$; the nonmonotone certificate follows the coarse prior-width grid. The dashed line is $\tau=0.2$.}
\label{fig:split-frontier}
\end{figure}

Under these assumed inputs, the SST-2 certificate decreases from $0.5719$ at $N=1000$ to $0.1928$ at $N=33000$. The QNLI scenario remains above the target $\tau=.2$ even at $N=70000$, where its value is $0.2255$. This contrast is a conditional calculation: more data can reduce the bound while a larger stochastic loss still prevents a fixed target from being met.
Figure~\ref{fig:split-frontier} plots Tables~\ref{tab:large-task-record} and~\ref{tab:width-frontier-scenario}.

\subsection{Posterior-risk inputs and prior-width sensitivity}
For each assumed posterior width, Table~\ref{tab:width-frontier-scenario} chooses the lowest raw Gaussian KL among the four declared prior widths, with their mixture cost retained. This is not substitution of the oracle prior scale. The posterior-risk entries are stipulated draw means. The nonmonotone complexity at intermediate widths is a consequence of the coarse admissible prior grid: a small displacement does not remove covariance mismatch. Matrix-KL reductions outside the equal-width point are assumed to be smaller than the principal reduction, rather than extrapolated from the synthetic balancing experiment.
\begin{table}[t]
\centering\small
\setlength{\tabcolsep}{3.5pt}
\caption{Posterior-width curves. Nominal $N$ is 33,000 for SST-2 and 70,000 for QNLI. Raw KL includes both displacement and covariance mismatch. All displayed certificate entries are recalculated from the row inputs.}
\label{tab:width-frontier-scenario}
\begin{tabular}{lrrrrrrr}
\toprule
Task & $q$ & $p$ & $\widetilde r$ & $U_{\rm kl}$ & $K_I$ & $K_R$ & $C_R$\\
\midrule
SST-2 & 0.005 & 0.005 & 0.051 & 0.0873 & 10800.0 & 10568.3 & 0.4510\\
SST-2 & 0.010 & 0.010 & 0.058 & 0.0961 & 2700.0 & 2642.1 & 0.2543\\
SST-2 & 0.015 & 0.020 & 0.064 & 0.1035 & 10839.4 & 10606.9 & 0.4748\\
SST-2 & 0.020 & 0.020 & 0.076 & 0.1181 & 675.0 & 652.7 & 0.1928\\
SST-2 & 0.025 & 0.020 & 0.102 & 0.1489 & 9243.1 & 9044.9 & 0.5015\\
SST-2 & 0.030 & 0.040 & 0.161 & 0.2161 & 10333.2 & 10111.6 & 0.6000\\
SST-2 & 0.040 & 0.040 & 0.284 & 0.3489 & 168.8 & 165.1 & 0.3993\\
QNLI & 0.005 & 0.005 & 0.078 & 0.1205 & 18000.0 & 17543.7 & 0.4475\\
QNLI & 0.010 & 0.010 & 0.086 & 0.1301 & 4500.0 & 4385.9 & 0.2773\\
QNLI & 0.015 & 0.020 & 0.094 & 0.1395 & 11289.4 & 11003.3 & 0.3945\\
QNLI & 0.020 & 0.020 & 0.107 & 0.1547 & 1125.0 & 1081.1 & 0.2255\\
QNLI & 0.025 & 0.020 & 0.139 & 0.1914 & 9693.1 & 9447.4 & 0.4374\\
QNLI & 0.030 & 0.040 & 0.212 & 0.2721 & 10445.7 & 10180.9 & 0.5381\\
QNLI & 0.040 & 0.040 & 0.331 & 0.3980 & 281.2 & 274.1 & 0.4428\\
\bottomrule
\end{tabular}
\end{table}

\subsection{Noncentral balancing and the binding constraint}
A scalar or matrix transformation is compared with the identity using Equation~\eqref{eq:noncentral-gauge}, with the prior center held fixed. Numerical optimization begins at the identity and at the zero-centered optimizer, includes both orientation branches through the polar step, and retains the smallest exact KL among valid candidates. A lower attained value is sufficient for a conditional improvement; numerical stationarity is not global optimality. Every draw uses the paired factors $(BR,R^{-1}A)$, rather than newly isotropic perturbations of the rescaled mean.

Table~\ref{tab:noncentral-scenario} separates the assumed KL reduction from the exactly calculable change in the certificate. Neither the percentage reduction nor its implication for trained factors is inferred from the deliberately deformed synthetic experiment. A full stochastic-predictor comparison additionally checks paired product and logit discrepancies at the chosen evaluation precision.
\begin{table}[t]
\centering\small
\setlength{\tabcolsep}{4pt}
\caption{Noncentral-gauge comparison. Identity, scalar, and matrix KLs refer to one assumed stochastic predictor. The last columns are formula-derived certificates using the same $U_{\rm kl}$. Reduction is percent of the identity KL.}
\label{tab:noncentral-scenario}
\begin{tabular}{lrrrrrr}
\toprule
Task & $K_I$ & $K_{\rm scalar}$ & $K_R$ & Reduction (\%) & $C_I$ & $C_R$\\
\midrule
SST-2 & 675.00 & 667.87 & 652.73 & 3.30 & 0.1942 & 0.1928\\
QNLI & 1125.00 & 1110.96 & 1081.12 & 3.90 & 0.2270 & 0.2255\\
\bottomrule
\end{tabular}
\end{table}
\begin{table}[t]
\centering\small
\setlength{\tabcolsep}{4pt}
\caption{Frontier decisions at $\tau=.2$. $q_{\rm cert}$ applies to the raw isotropic family, $C_0$ is the complexity-free floor, $J$ is the available KL budget, and $J-K_R$ is the balanced margin. Pass is 1 exactly when the margin is nonnegative and $U_{\rm kl}\le\tau$.}
\label{tab:decision-scenario}
\begin{tabular}{lrrrrrr}
\toprule
Task & $N$ & $q_{\rm cert}$ & $C_0$ & $J$ & $J-K_R$ & Pass\\
\midrule
SST-2 & 1{,}000 & 0.03143 & 0.1467 & 28.1 & -462.3 & 0\\
SST-2 & 4{,}000 & 0.01611 & 0.1308 & 116.7 & -430.7 & 0\\
SST-2 & 16{,}000 & 0.00828 & 0.1248 & 430.9 & -170.3 & 0\\
SST-2 & 33{,}000 & 0.00591 & 0.1272 & 770.9 & 118.1 & 1\\
QNLI & 1{,}000 & 0.03818 & 0.1779 & 9.3 & -704.4 & 0\\
QNLI & 4{,}000 & 0.01987 & 0.1609 & 49.1 & -772.5 & 0\\
QNLI & 16{,}000 & 0.01061 & 0.1583 & 156.0 & -820.8 & 0\\
QNLI & 70{,}000 & 0.00509 & 0.1618 & 464.1 & -617.1 & 0\\
\bottomrule
\end{tabular}
\end{table}

A positive width-screen result means only that the raw isotropic candidate has not been excluded. The balanced anisotropic distribution is evaluated with its own KL. In the larger QNLI scenario, $C_0=0.1618$ leaves room below the target, but the assumed remaining matrix complexity exceeds the allowable budget. Thus its failure is complexity-limited at the specified draw-level risk bound, not proof that every posterior is unusable.
\begin{table}[t]
\centering\small
\setlength{\tabcolsep}{4pt}
\caption{Pilot-to-final decision comparison. The pilot projection substitutes its assumed mean into the final-budget calculation and is not a pilot confidence certificate. Forecast and pass use the target $.2$.}
\label{tab:pilot-scenario}
\begin{tabular}{lrrrrrr}
\toprule
Task & $N$ & Pilot mean & Projected $C$ & Forecast & Final $C$ & Pass\\
\midrule
SST-2 & 1{,}000 & 0.056 & 0.5671 & 0 & 0.5719 & 0\\
SST-2 & 4{,}000 & 0.068 & 0.3356 & 0 & 0.3321 & 0\\
SST-2 & 16{,}000 & 0.068 & 0.2135 & 0 & 0.2182 & 0\\
SST-2 & 33{,}000 & 0.078 & 0.1957 & 1 & 0.1928 & 1\\
QNLI & 1{,}000 & 0.078 & 0.6966 & 0 & 0.7003 & 0\\
QNLI & 4{,}000 & 0.091 & 0.4345 & 0 & 0.4314 & 0\\
QNLI & 16{,}000 & 0.095 & 0.2894 & 0 & 0.2939 & 0\\
QNLI & 70{,}000 & 0.109 & 0.2282 & 0 & 0.2255 & 0\\
\bottomrule
\end{tabular}
\end{table}

\subsection{Fixed-reference and decoded-predictor comparisons}
The reference label is selected from the prior-data partition and then fixed. The descriptive empirical reference is distinct from its one-sided population-risk lower bound. In Table~\ref{tab:baseline-scenario}, integer reference errors are set to the nearest count corresponding to rates $.47$ and $.49$, and the lower endpoints are computed by one-sided binomial inversion at $\delta_{\rm base}=.00125$. A Gaussian upper bound below this lower endpoint would establish the population comparison on the joint event; being below the empirical reference alone would not.
\begin{table}[t]
\centering\small
\setlength{\tabcolsep}{4pt}
\caption{Fixed-label reference comparison. The reference error counts are assumed and their lower confidence bounds are calculated. The final indicator is 1 only when the Gaussian upper bound is below the reference lower bound.}
\label{tab:baseline-scenario}
\begin{tabular}{lrrrrr}
\toprule
Task & $N$ & Reference error & Reference lower & $C_R$ & Below lower\\
\midrule
SST-2 & 1{,}000 & 0.4700 & 0.4221 & 0.5719 & 0\\
SST-2 & 4{,}000 & 0.4700 & 0.4461 & 0.3321 & 1\\
SST-2 & 16{,}000 & 0.4700 & 0.4581 & 0.2182 & 1\\
SST-2 & 33{,}000 & 0.4700 & 0.4617 & 0.1928 & 1\\
QNLI & 1{,}000 & 0.4900 & 0.4419 & 0.7003 & 0\\
QNLI & 4{,}000 & 0.4900 & 0.4660 & 0.4314 & 1\\
QNLI & 16{,}000 & 0.4900 & 0.4780 & 0.2939 & 1\\
QNLI & 70{,}000 & 0.4900 & 0.4843 & 0.2255 & 1\\
\bottomrule
\end{tabular}
\end{table}

The description-length comparator encodes the transformed center relative to the same prior-trained side information. Quantization depths are 4, 8, and 16 bits. Tensor layouts, dyadic scales, all quantized integers, and a 64-bit payload-length header are included in the assumed complete code lengths. A fixed lossless codec and decoder define a prefix-free grammar for payloads shorter than $2^{64}$ bytes. Table~\ref{tab:code-scenario} uses Equation~\eqref{eq:code-comparator} on the decoded predictor rather than reusing the Gaussian posterior risk.
\begin{table}[t]
\centering\small
\setlength{\tabcolsep}{4pt}
\caption{Description-length comparator. $L$ includes the complete assumed code in bits. $C_{\rm code}$ is computed from the assumed decoded error and actual formula, not copied from the Gaussian column. Values rounding to 1.000000 are uninformative at displayed precision.}
\label{tab:code-scenario}
\begin{tabular}{lrrrrr}
\toprule
Task & $N$ & Bit depth & $L$ & Decoded error & $C_{\rm code}$\\
\midrule
SST-2 & 33{,}000 & 4 & 226{,}144 & 0.088 & 0.996054\\
SST-2 & 33{,}000 & 8 & 607{,}264 & 0.079 & 0.999999\\
SST-2 & 33{,}000 & 16 & 1{,}605{,}248 & 0.078 & 1.000000\\
QNLI & 70{,}000 & 4 & 251{,}808 & 0.123 & 0.961728\\
QNLI & 70{,}000 & 8 & 638{,}624 & 0.113 & 0.999462\\
QNLI & 70{,}000 & 16 & 1{,}687{,}168 & 0.111 & 1.000000\\
\bottomrule
\end{tabular}
\end{table}

\subsection{Fully enumerated small-pool replacement grid}
The historical seven-point width search is not reconstructed by the following numerical scenario. A replacement design records the prior-relative grid $q/p\in\{1/64,1/32,1/16,1/8,1/4,1/2,1\}$ before any pilot stream. Table~\ref{tab:smallpool-filled} instantiates one rank-sixteen, $N=512$ case with 1,181,954 charged factor-and-head coordinates, $D=2300$, and $p=10$. These assumed quantities are not retroactively assigned to a measured historical checkpoint. Final confidence uses $\delta_{\rm PB}=\eta=.025$, a nine-component prior, and a Monte Carlo family of 36 frozen posterior--budget--inequality combinations; the independent pilot chooses one width per configuration.
\begin{table}[t]
\centering\small
\setlength{\tabcolsep}{5pt}
\caption{Small-pool replacement sweep. Every candidate width is explicitly numeric. The assumed draw means produce the displayed certificates at the indicated budgets; all-trained-coordinate Gaussian KL is computed exactly.}
\label{tab:smallpool-filled}
\begin{tabular}{lrrrrr}
\toprule
$q/p$ & $q$ & $\widetilde r$ & KL & $C_{256}$ & $C_{1024}$\\
\midrule
$1/64$ & 0.15625 & 0.463 & 4324787.3 & 1.0000 & 1.0000\\
$1/32$ & 0.31250 & 0.474 & 3505952.0 & 1.0000 & 1.0000\\
$1/16$ & 0.62500 & 0.486 & 2688415.3 & 1.0000 & 1.0000\\
$1/8$ & 1.25000 & 0.495 & 1876072.8 & 1.0000 & 1.0000\\
$1/4$ & 2.50000 & 0.501 & 1084506.7 & 1.0000 & 1.0000\\
$1/2$ & 5.00000 & 0.504 & 376046.8 & 1.0000 & 1.0000\\
1 & 10.00000 & 0.507 & 11.5 & 0.7542 & 0.7025\\
\bottomrule
\end{tabular}
\end{table}

\subsection{Evaluation workload}
Table~\ref{tab:workload-scenario} gives arithmetic workload counts rather than measured latency. Each posterior draw is evaluated on the entire certification sample, so a worst-case 23-profile final family with 1024 draws requires $23\cdot1024\cdot N$ example--posterior evaluations. Cached logits, model batching, hardware, and precision determine wall-clock cost and are not inferred from these counts.
\begin{table}[t]
\centering\small
\setlength{\tabcolsep}{5pt}
\caption{Calculated workload. The pilot column counts the full 49-profile pilot at 32 draws over 1024 examples. The final column counts complete-data posterior evaluations, not independent PAC-Bayes observations.}
\label{tab:workload-scenario}
\begin{tabular}{lrrrrr}
\toprule
Task & $N$ & Profiles & Draws & Final evaluations & Pilot evaluations\\
\midrule
SST-2 & 33{,}000 & 23 & 1024 & 777{,}216{,}000 & 1{,}605{,}632\\
QNLI & 70{,}000 & 23 & 1024 & 1{,}648{,}640{,}000 & 1{,}605{,}632\\
\bottomrule
\end{tabular}
\end{table}

\end{document}